\documentclass[final, 5p, times, twocolumn]{elsarticle}

\usepackage[ruled,vlined]{algorithm2e}
\usepackage{amssymb}
\usepackage{amsmath}
\usepackage{pgfplots}
	\pgfplotsset{compat=1.11}
\usepackage{graphicx}
\usepackage{subcaption}
\usepackage{booktabs}
\usepackage{epstopdf}
\usepackage{array}
\usepackage[utf8]{inputenc}
\usepackage{wrapfig}
\usepackage[pdftex,
		    bookmarks={true},
		    bookmarksopenlevel={1},
		    bookmarksnumbered={true}]   
	{hyperref}
	
\makeatletter
\def\BState{\State\hskip-\ALG@thistlm}
\makeatother

\journal{Applied Soft Computing}
\begin{document}
\begin{frontmatter}


\title{Change points detection in crime-related time series: an 
on-line fuzzy approach based on a shape space representation} 
\author[imi]{Fabrizio Albertetti\corref{fa_cor}}
\ead{fabrizio.albertetti@unine.ch}
\author[ips]{Lionel Grossrieder}
\ead{lionel.grossrieder@unil.ch}
\author[ips]{Olivier Ribaux}
\ead{olivier.ribaux@unil.ch}
\author[imi]{Kilian Stoffel}
\ead{kilian.stoffel@unine.ch}
\address[imi]{Information Management Institute, University of Neuchatel, 
CH-2000 Neuchatel, Switzerland} 
\address[ips]{School of Criminal Sciences, University of Lausanne, CH-1015 
Lausanne-Dorigny, Switzerland}
\cortext[fa_cor]{Corresponding author. Tel.: +41 32 718 14 49}

\begin{abstract}
The extension of traditional data mining methods to time series has 
been effectively applied to a wide range of domains such as finance, 
econometrics, biology, security, and medicine. Many 
existing mining methods deal with the task of change points detection, but very 
few provide a flexible approach. Querying specific change points 
with linguistic variables is particularly useful in crime 
analysis, where intuitive, understandable, and appropriate detection of changes 
can significantly improve the allocation of resources for timely and concise 
operations. In this paper, we propose an on-line method for detecting and 
querying change points in crime-related time series with the use of a 
meaningful representation and a fuzzy inference system. Change points detection 
is based on a shape space representation, and linguistic terms describing 
geometric properties of the change points are used to express queries, offering 
the advantage of intuitiveness and flexibility. An empirical evaluation is  
first conducted on a crime data set to confirm the validity of the 
proposed method and then on a financial data set to test its general 
applicability. A comparison to a similar change-point detection algorithm and a 
sensitivity analysis are also conducted. Results show that the method is able 
to accurately detect change points at very low computational costs. 
More broadly, the detection of specific change points within time 
series of virtually any domain is made more intuitive and more understandable, 
even for experts not related to data mining.
\end{abstract}

\begin{keyword}
change points detection \sep
qualitative description of data \sep
time series analysis \sep
fuzzy logic \sep
crime analysis
\end{keyword}

\end{frontmatter}


\section{Introduction}
\label{sec:introduction}
The analysis of time series naturally arises in crime analysis as well as in 
any data-driven domain. Finding sudden changes in criminal activities is a 
particular task known as change points detection. In this paper, a flexible 
on-line change points detection method for helping crime 
analysts to easily and understandably monitor changes is proposed. Change 
points are detected in two steps: the segmentation of the time 
series and the querying of points with a fuzzy inference system.

\subsection{Motivation}
Knowledge extraction of time series can be viewed as an extension of 
traditional mining methods with an emphasis on the temporal aspect. Among 
these, \textit{Change points detection} methods focus on finding time points at 
which data \textit{suddenly} change (in contrast to \textit{slow} changes).
Many studies have shown interesting applications of change points detection 
in various domains. These methods are based on neural networks, regressions, or 
other statistical models, with an emphasis on the efficiency of these 
methods. However, only a few consider approaches with these two properties: a 
meaningful and expressive subspace representation of the time series, and a 
dynamic segmentation process without fixed-sized windows, linked together 
flexibly.

In the domain of crime analysis, such flexible and intuitive approaches for 
change points detection are particularly sought, especially for crime trends 
monitoring. Previous studies from the authors (\cite{Albertetti2012}, 
\cite{Albertetti2013}, \cite{Albertetti2013b}, and 
\cite{Grossrieder2013}) emphasize on the usefulness of crime trends monitoring 
activities and advocate the use of appropriate methods for considering the 
specificities and the constraints of the crime analysis domain, that is 
basically dealing with uncertainties. The automated process of 
change points detection is considered as a major step in the production of 
intelligence, supporting the activity of crime analysis (also sometimes 
referred to crime intelligence).

Flexible change points detection methods are critical for supporting analysts 
in their daily tasks, especially for the monitoring of serial and high-volume 
crimes (e.g., burglaries). Most of the time, crime analysts have no particular 
background in time series analysis, but still need to analyze and monitor crime 
trends. These trends are drawn into the whole activities and are not always 
perceived by police forces. As for example, querying criminal activities about 
a particular increase in crime trends for targeted police interventions, as 
well as querying patterns of changes for the general understanding of crime 
phenomena are common tasks.

Finding changes in crime trends assumes two conditions: (a) the actual 
existence of a trend, and (b) its detection within the data. The first 
condition is far from obvious, but as crime analysis is founded on 
environmental criminology theories, a justification for the existence of crime 
trends appears (\cite{Grossrieder2013}, \cite{Boba2009}, and 
\cite{Felson1998}). The second is generally simply assumed, but difficult 
to detect in massive data sets and needs intuitive and understandable 
analytical methods.

Although the proposed method is a specific answer for the domain of crime 
analysis, we believe that it has great potential applications 
in several domains. As an example, in the financial domain, it proves very 
useful to find and query change points in real-time, giving the investors 
flexible means to detect trends indicating the right moment for selling or 
buying stocks.

\subsection{Contribution of this paper}
The method proposed in this paper, for the Fuzzy Change Points Detection in 
crime-related time series (FCPD), aims to focus 
on flexibility and intuitiveness. To achieve this purpose, the method combines 
a segmentation step and a querying step. Moreover, the following 
characteristics make 
FCPD unique:

	\begin{itemize}
	\item a meaningful and expressive representation of the time series is used 
	;
	\item the segmentation is \textit{dynamic}, that is, segments are set 
	according to the underlying shapes of the time series, without using a 
	fixed-size window parameter;
	\item changes are queried with linguistic terms, using a fuzzy inference 
	system;
	\item the method does not rely on training sets;
	\item the method is on-line and iterative, i.e., change points can be 
	detected with past values only and there is no need to compute the entire 
	model at each new observed value. The computational cost is very low; it is
	related to the size of the approximating polynomials (instead of the number 
	of the observations).
	\end{itemize}
	
Indeed, with the use of a meaningful representation and a dynamic 
segmentation, change points can be more easily described and identified. The 
segments found in a time series, reflecting change points, are 
described with meaningful estimators such as the average, the slope, the 
curvature, etc. Then, with the use of a fuzzy inference system, a query can be 
specified using linguistic terms describing the geometric estimators. It 
becomes then easier, for instance, to query a time series about the most abrupt 
changes in terms of slope. In the example ``IF \textit{average} 
is \textit{low} AND \textit{slope} is \textit{very\_high}, 
THEN \textit{pertinence} is HIGH'', the inference system would return a high 
score on segments representing shapes of the given description. This approach 
makes the querying of change points particularly intuitive and flexible, 
especially for domain experts.

\subsection{Structure of this paper}
The remainder of this paper is structured as follows: in Section 
\ref{sec:literature}, a literature review in the mining of time series is 
provided; Sect. \ref{sec:concepts} introduces some concepts in the preparation, 
representation, and analysis of time series; Sect. \ref{sec:proposed} details 
FCPD, a step-by-step method for the fuzzy querying and detection of change 
points in crime-related time series; an 
empirical validation on synthetic and real-world data is conducted in Sect. 
\ref{sec:experiments}; results are discussed in Sect.\ref{sec:discussion}; and 
finally in Sect. \ref{sec:conclusion} a conclusion is drawn from the 
experiments and some tracks for future work are suggested.

\section{Literature review}
\label{sec:literature}
Change points detection has numerous application domains, 
as for example finance, biology, ocean engineering, medicine, and crime 
analysis. It is considered as a final objective in the 
whole process of time series analysis among classification, rules discovery, 
prediction, and summarization. Almost all of these mining tasks require data 
preparation, namely the representation of the time series, its indexing, its 
segmentation, and/or its visualization. 
In this section, we propose a review of these steps, before comparing existing 
methods for change points detection. An extensive review of the analysis of 
time series can be found in \cite{Fu2011}, as well as a general methodology in 
\cite{Last2001}.

\subsection{Representation of time series}
Many representation models of time series have been dealt with in the 
literature, each claimed with relative advantages and drawbacks. Two main 
categories are symbolic representations and numeric representations. Symbolic 
representations are less sensitive to noise and are usually computationally 
faster. For the last decade, the community has been paying particular attention 
to the Symbolic Aggregate Approximation (SAX) representation (\cite{Lin2003} 
and \cite{Lin2007}), with the main advantages to reduce the original 
dimensionality of the data, being on-line, and having a robust distance 
measure. However, it does not cover all needs. In \cite{Fuchs2010b}, a numeric 
representation ---which differs from SAX and many others by giving a meaning to 
the representation--- is used to perform several mining tasks. This 
\textit{shape space} representation uses coefficients as shape estimators of 
the time series it represents, leading to an intuitive description.

\subsection{Segmentation of time series}
Most mining methods use subsequences (or segments) of time series as input to 
the analysis. Segmentation algorithms with the approach of a sliding window are 
simple to use but present the main drawback of being static, i.e., 
segmenting the time series according to a fixed and exogenous parameter (e.g., 
the length of the window) without considering the observed values. Other 
algorithms, based on a bottom-up or a top-down approach are considered as 
dynamic (e.g., by using some error criteria as segmentation thresholds) but 
need the whole data set to operate. These off-line algorithms usually perform 
better in terms of accuracy but have higher computational costs and are not 
suitable for real-time applications. A combination of the aforementioned 
algorithms, namely the SWAB segmentation algorithm, is presented in 
\cite{Keogh2001}. A study \cite{Keogh2004} provides benchmarks on these claims 
and as a result suggests that SWAB is empirically superior to all other 
algorithms discussed in the literature. As we believe there is no 
silver bullet, each application has its own requirements. A more flexible 
approach is the SwiftSeg algorithm \cite{Fuchs2010}, providing a 
dynamic and on-line approach to segmentation, with the possibility of a 
mix between growing and sliding window. Another interesting segmentation 
approach \cite{Chung2002}, specific to stock mining and described as 
dynamic, is based on the identification of perceptually important points 
(PIP).

\subsection{Fuzzy analysis of time series}
A small subset of temporal mining methods takes advantage of the 
characteristics offered by fuzzy logic and fuzzy sets. The concept of fuzzy 
time series has first been defined by Song and Chissom in \cite{Song1993} and 
\cite{Song1993b}, with an application in class enrollment forecasting. Soon 
followed multiple variations and improvements of the basic method (e.g., 
\cite{Chen1996}, \cite{Hwang1998}, \cite{Huarng2001}, or \cite{Chen2012}), with 
their own types of fuzzy inference systems (FIS). Two common FIS, namely the 
Mamdani inference system \cite{Mamdani1974} and the Takagi-Sugeno inference 
system \cite{Sugeno1985}, can be intuitively used to deal with uncertain and 
flexible data. In \cite{Lee2006}, an application in finance uses 
an FIS for pattern discovery. In \cite{Guener2014}, prediction of long shore 
sediments is also dealt with the use of an FIS. In parallel, a combination of 
FISs and neural networks have found an origin in \cite{Jang1993}. As for 
examples, the prediction of time series is performed with dynamic evolving 
neuro-fuzzy inference systems \cite{Song2000}, the classification of 
electroencephalograms \cite{Gueler2005}, as well as the prediction of 
hydrological time series \cite{Zounemat-Kermani2008}.

\subsection{Change points detection}
Change points detection in time series analysis has been 
thoroughly investigated, mainly using statistical models 
(see \cite{Basseville1993} for a general introduction). Reeves et al. 
\cite{Reeves2007} attempt to review and compare the major change points 
detection methods for climate data series.

More specifically, related approaches for change points detection have been 
investigated in a relatively limited set of studies. For example, a statistical 
based approach using fuzzy clustering is described in \cite{Wu1999,Kumar2001}. 
Verbesselt et al., in \cite{Verbesselt2010} and \cite{Verbesselt2010b}, detect 
breaks for additive seasonal and trends (BFAST), with a principal application 
is phenology. To deal with imprecise observation in time series, changes are 
analyzed with fuzzy variables in \cite{Cappelli2013}. In \cite{Chen2013}, a 
contextual change detection algorithm addresses relative changes with respect 
to a group of time series. In \cite{Yamanishi2002} and \cite{Takeuchi2006}, the 
utility of a framework for outliers detection of time series prediction is 
highlighted. In \cite{Chen2012}, the need to use linguistic values for 
comprehensible results is advocated, where fuzzy time series mining is used for 
association rules between data points (but not between segments) with 
fixed-size window. A qualitative description of multivariate time series with 
the use of fuzzy logic is presented in \cite{Moreno-Garcia2014}. Yu et al. 
\cite{Yu2001} propose a fuzzy piecewise regression analysis with automatic 
change points detection. In \cite{Wang2004b}, DoS attacks are monitored with a 
change point approach based on the non-parametric Cumulative Sum (CUSUM) method.


\section{Time series representation and fuzzy concepts}
\label{sec:concepts}
In the following subsections, a review of a time series representation using 
polynomials is presented and their main advantages are explained. Then a 
dynamic segmentation method is described. Finally, concepts of fuzzy time 
series and fuzzy inference systems, which are useful to analyze segments, are 
introduced.

\subsection{A polynomial shape space representation}
\label{ssec:representation}
Let us consider a time series defined by the sequence

\begin{equation}\label{eq:s}
	s=(y_0,y_1,\ldots,y_N), \quad y_i \in \mathbb{R} \quad (i=0,1,\ldots,N)
\end{equation}

\noindent of $N+1$ points measured over the equidistant points in time 
$x_0,x_1,\ldots,x_N$. 
Basically, our set of points $s$ can be modeled by a parametrized function 
$f(x)$, which is obtained with a 
linear combination of basis functions $f_k$:

\begin{equation}\label{eq:lc}
	f(x)=\sum_{k=0}^{K}w_k \: f_k(x) \,.
\end{equation} 

\noindent The properties of this approximation depend on the choice of the 
basis functions $f_k$ and their weights. Given some appropriate basis 
functions, an optimal approximation 
can be found with the vector of weights $\mathbf{w^{\ast}}\in 
\mathbb{R}^{K+1},\mathbf{w^{\ast}}=(w_0,w_1,\ldots,w_k)^T$, which minimizes the 
approximation error in the least-squared sense. Fuchs et 
al., in \cite{Fuchs2010}, claim that these weights show 
interesting properties when using some specific of these $K+1$ basis functions. 
Indeed, when particular conditions are met, these weights describe 
the shape of the considered time series intuitively. As a corollary, 
an efficient similarity measure can be defined based on the extracted features.

Let us now describe these particular approximating polynomials, as in 
\cite{Fuchs2010}, with

\begin{equation}\label{eq:orth-exp}
	p(x)=\sum_{k=0}^{K} \alpha_k \: p_k(x) \,,
\end{equation}

\noindent where $p(x)$ is the approximating polynomial, the polynomials $p_k$ 
are the 
basis functions $f_k$ and the coefficients $\alpha_k$ are the weights $w_k$, 
relating to Equation \ref{eq:lc}. These coefficients are defined as

\begin{equation}
	\alpha_k = \frac{1}{\|p_k\|^2}\sum_{n=0}^{N}y_n \: p_k(n) \,,
\end{equation}

\noindent where

\begin{flalign*}
	p_{-1}(x)&=0 \,, \\
	p_{0}(x)&=1 \,, \\
	p_{k+1}(x)&=(x-a_k) \: p_k(x)-b_k \: p_{k-1}(x) \,.
\end{flalign*}

\noindent Then, by defining $\boldsymbol{\alpha}$ as the \textit{vector of 
	coefficients}

\begin{equation}
	\boldsymbol{\alpha}=\left(\alpha_0, \alpha_1, \ldots, \alpha_K \right)^T \,,
\end{equation}

\noindent any time series can be characterized by these coefficients only, 
given some polynomials. The interesting property is that the $i_{th}$ 
coefficient represents the $i_{th}$ derivative of the approximated time series; 
i.e., the first coefficient $\alpha_0$ is an optimal estimator (in the 
least-squared sense) for the average of the considered $N+1$ data points, 
$\alpha_1$ an estimator for the slope, $\alpha_2$ an estimator for the 
curvature, etc.

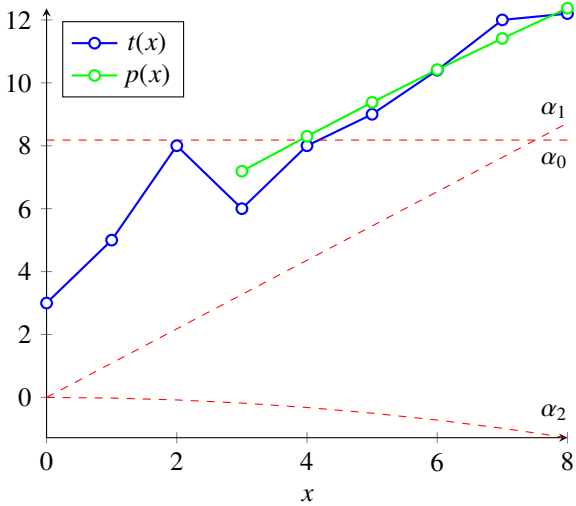
\begin{figure}[htp]
	\centering
	\begin{tikzpicture}
	\begin{axis}[axis lines = left, xlabel = $x$, legend pos=north 
	west]	
	\addplot[color=blue,solid,thick,mark=*, mark options={fill=white}] 
	coordinates {
		(0,3)(1,5)(2,8)(3,6)(4,8)(5,9)(6,10.4)(7,12)(8,12.2) 
	};
	\addlegendentry{$t(x)$}
	\addplot[color=green,solid,thick,mark=*, mark options={fill=white}] 
	coordinates {
		(3,7.19)(4,8.30)(5,9.38)(6,10.42)(7,11.41)(8,12.37) 
	};	
	\addlegendentry{$p(x)$}
	\addplot[domain=0:8, samples=9, color=red, dashed]
	{8.18}; 
	\addplot[domain=0:8, samples=9,	color=red, dashed]
	{1.09*x}; 
	\addplot[domain=0:8, samples=9,	color=red, dashed]
	{-0.02*x^2}; 
	\node[below] at (axis cs:  7.8, 8.1) {$\alpha_0$};
	\node[above] at (axis cs:  7.8, 8.6) {$\alpha_1$};
	\node[above] at (axis cs:  7.8, -1) {$\alpha_2$};
	\end{axis}
	\end{tikzpicture}
	\caption[A time series and its polynomial approximation.]{A time series 
		$t(x)$ and its polynomial 
		approximation $p(x)$ with $K=2$, and their coefficients. The 
		coefficients $\alpha_0=8.18$ (estimator for the average), 
		$\alpha_1=1.09$ (estimator 
		for the slope), and	$\alpha_2=-0.02$ (estimator for the curvature) are 
		depicted with the first term only of their respective polynomials. It 
		should be noticed that the approximation can only start from $K+1$ data 
		points.}
	\label{fig:estimators} 
\end{figure}

The parameter $K$ of the orthogonal expansion (Eq. \ref{eq:orth-exp}) has 
to be carefully chosen in accordance with the desired description of the time 
series. As depicted in Figure \ref{fig:estimators}, setting $K=0$ defines a 
single polynomial term in Eq. 
\ref{eq:orth-exp} with a maximum degree of 0 and a corresponding vector 
coefficient $\boldsymbol{\alpha}\in \mathbb{R}^1$, representing a time 
series according to its average only; setting $K=1$ adds the estimator 
of slope; setting $K=2$ adds on top the estimator of curvature; and so on. 
Choosing this parameter is a trade-off between computational costs and  
representation accuracy.

In order to hold these desired properties, each of the given polynomials $p_k$ 
defining the \textit{orthogonal expansion} of the approximation must:

\begin{itemize}
	\item have different ascending degrees 0,1,...,K;
	\item have a leading coefficient of 1;
	\item be orthogonal with respect to the inner product
	\begin{equation}
		\langle p_{i} | p_{j} 
		\rangle=\sum_{n=0}^{N}p_{i}(x_n) \: p_{j}(x_n), \quad
		i\neq j \,.
	\end{equation}
\end{itemize}

For instance, the discrete Chebyshev's polynomials reveal these 
criteria. Defined in a recursive way, the first Chebyshev's terms are:

\begin{flalign*}
	p_0(x) =& 1, &\\
	p_1(x) =& x-\frac{N}{2} \,,\\
	p_2(x) =& x^2-Nx+\frac{N^2-N}{6} \,,\\
	p_3(x) =& x^3-\frac{3N}{2}x^2 + \frac{6N^2-3N+2}{10}x \\
	& -\frac{N(N-1)(N-2)}{20} \,.
\end{flalign*}

\noindent More generally, a term of the series is defined as 

\begin{equation} 
	p_{k+1}(x) = \left(x-\frac{N}{2}\right) \: p_k(x)
	-\frac{k^2((N+1)^2-k^2)}{4(4k^2-1)} \: p_{k-1}(x) \,,
\end{equation}

\noindent and their squared norms are given by

\begin{multline}
\|p_k\|^2=\frac{(k!)^4}{(2k)! \: (2k+1)!}\prod_{i=-k}^{k}(N+1+i) \,, \\
\quad k=0,1,\ldots,K \,.
\end{multline}	

\noindent Based on these definitions, it is now possible to redefine our time 
series $s$ 
from Eq. \ref{eq:s}. As the vector $\boldsymbol{\alpha}$ contains the 
estimators up to degree $K$ for the considered time series, it is said that  
\textit{$s$ is approximated by $\boldsymbol{\alpha}$} with the statement

\begin{equation}
	s=(y_0,y_1,\ldots,y_N)\sim \boldsymbol{\alpha} \,,
\end{equation}

where $s\in \mathbb{R}^{N+1},\quad  \boldsymbol{\alpha}\in 
\mathbb{R}^{K+1},\quad K\ll N$.

\subsection{Segmenting time series}
\label{ssec:seg}
Segmenting time series is useful for analyzing and comparing subsets of data 
points. The considered segmentation approach in this paper is 
\textit{dynamic}, 
meaning segmentations are performed in accordance with the intrinsic shapes 
underlying in the data, in contrast to other segmentation approaches that only 
depend on an artificial window size or with equal size segments. To do so, the 
sequence $s$ from Eq. \ref{eq:s} is split into the set of contiguous windows
 
	\begin{equation}
	W(s)=\bigcup_{s^{(i)}\in s}\: s^{(i)} \,, \quad i \in 
	\{0,1,\ldots,N/2\} \,,
	\end{equation}
	
\noindent where $W$ is a partition of $s$, and $s^{(i)}$ is the $i_{th}$ 
segment 
containing two or more elements. To be dynamic, the partitioning is done 
by detecting \textit{change points} $\boldsymbol{\hat{x}}$ within our 
time domain $x=(x_0, x_1, \ldots, x_N)$, relating to the concept of abrupt 
changes 
detection \cite{Basseville1993}.

Let us consider a small example. Within the sequence $s=(y_0, y_1, \ldots, 
y_{10})$, the change points $\hat{x}$ detected are $x_3$ 
and $x_7$; then $\boldsymbol{\hat{x}}=(3,7), \: 
W(s)=\{s^{(1)},s^{(2)},s^{(3)}\}$, 
with 
$s^{(1)}=(y_0,y_1,y_2,y_3), \: 
s^{(2)}=(y_4,y_5,y_6,y_7), \: 
s^{(3)}=(y_8,y_9,y_{10})$.
It has to be emphasized that these contiguous windows do not need to have the 
same size. In fact, their underlying estimators $\boldsymbol{\alpha}$ only 
depend on the shape of the segment. Therefore, \textit{primitives}, or 
\textit{basic} shapes are more accurately represented by these estimators, as 
the deviation of the predicted values (the sum of the residuals) is 
low within the considered segment (i.e., the estimators do not significantly 
change within a segment 
while the windows is growing). 
This property justifies why an adaptive way of segmenting the time series with 
change points is preferred to fixed-size window segmentation methods, 
considering the objective of this study. 

	\begin{figure*}[htp]
	\captionsetup[subfigure]{position=b}
	\centering
	\caption{Illustration of the segmentation of a time series with its 
	shape-space	representation.}
	\label{fig:segmentation}
	\subcaptionbox{A real time series of 72 samples. The time series has been 
	normalized ($\mu=0$, $\sigma^2=1$). From the pretty chaotic shape of the 
	series, many change points are supposed to be found.
		\label{fig:segmentation_a}}
		{\includegraphics[width=.3\linewidth]{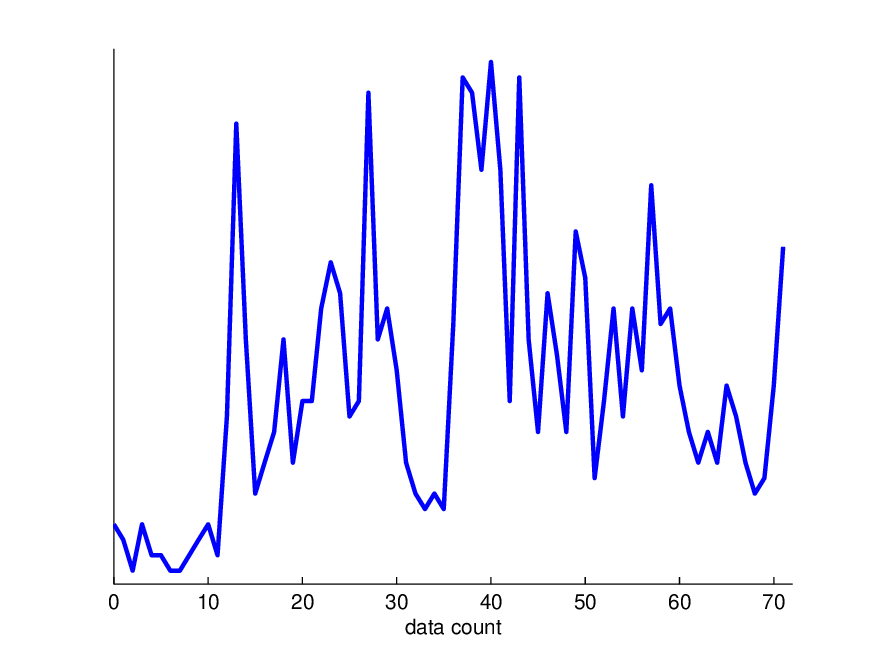}}
		\hfill
	\subcaptionbox{Segmentation of the time series, with $K=3$. Twelve 
	different segments are found (represented by vertical lines). The 
	segmentation here is based on the number of sign changes of the slope and 
	the deviation of the predicted value. 
		\label{fig:segmentation_b}}
		{\includegraphics[width=.3\linewidth]{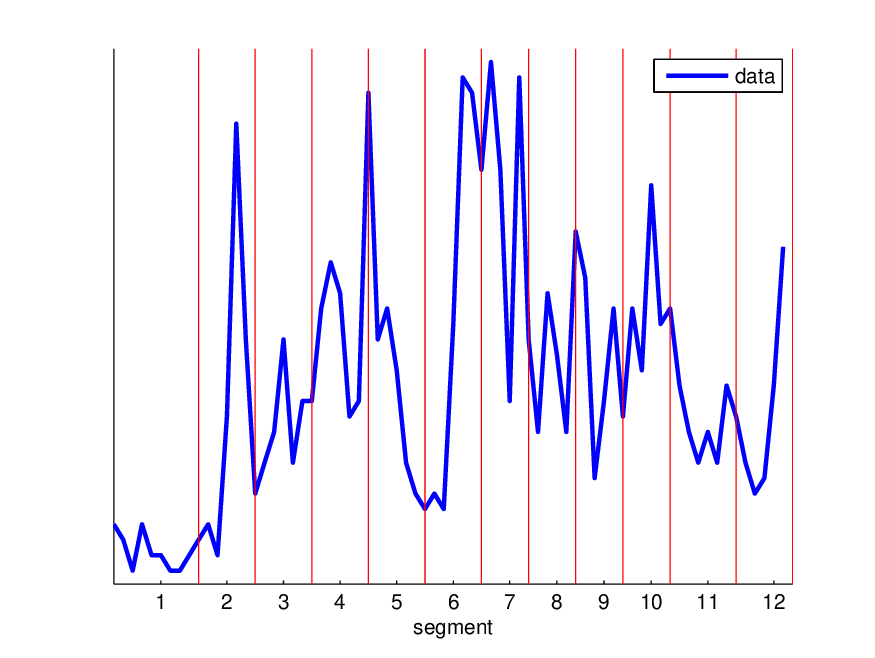}}
		\hfill
	\subcaptionbox{Segments depicted with their shape space representation 
	(only the first two coefficients are shown). The segments 1 and 7 are 
	easily 
	identified as with the lowest and highest average ($\alpha_0$) and the 
	segments 5 and 6 with the lowest and highest slope ($\alpha_1$).
		\label{fig:segmentation_c}}
		{\includegraphics[width=.3\linewidth]{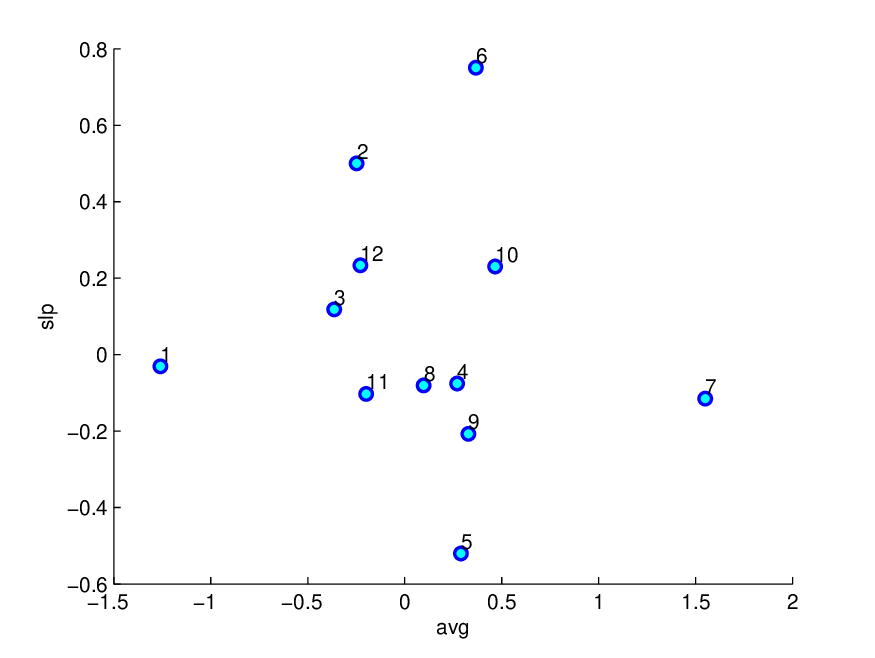}}
	\end{figure*}
	
As part of an iterative process, the first segmentation step starts with a 
window from the first point of the segment, setting $x=0$. The corresponding  
orthogonal expansion is computed, holding the $\boldsymbol{\alpha}$ 
coefficients. Then, specific criteria based on the coefficients are derived 
(such as the deviation of the predicted value or the count of sign switches of 
the slope) and compared to thresholds. If the thresholds are exceeded, the 
growing process is stopped, a change point is detected and a new segment starts 
with the next available data point; otherwise, the window keeps growing to the 
next point, the expansion is updated and the new coefficients are again 
compared. Figure \ref{fig:segmentation} depicts an example of this segmentation.

This segmentation method is based on the \textit{SwiftSeg} algorithm from Fuchs 
et al. \cite{Fuchs2010}. Their study describes an on-line algorithm for 
updating the values of the coefficients, where the computation only depends
on the last point added to the window, leading to effective computational 
costs; in contrast to off-line algorithms that need the entire window to update 
the coefficients. Combinations of growing and fixed-length window are also 
documented and experimented.

\subsection{Fuzzy time series}
\label{ssec:fts}
The concept of fuzzy time series as first defined by Song and Chissom in 
\cite{Song1993} is here resumed. Let us consider the universe of discourse

	\begin{equation}
	U=\{u_1,u_2,\ldots,u_m\}
	\end{equation}
	
\noindent and the set

	\begin{equation}
	A_i=\mu_{A_i}(u_1)/u_1+\ldots+\mu_{A_i}(u_m)/u_m \,.
	\end{equation}
	
\noindent $A_i$ is a fuzzy set of $U$, where '/' indicates the separation 
between the 
membership grades and the elements of the universe of discourse $U$, '+' is the 
union of two elements, and the fuzzy membership function

	\begin{equation}
	\mu_{A_i}(u_j):U \to [0,1]
	\end{equation} 
	
\noindent expresses the grade of membership of $u_j$ in $A_i$.

Let the elements of our time series $(y_t)(t=0,1,\ldots,N)$, a subset of 
$\mathbb{R}$, be the universe of discourse replacing $U$ on which the fuzzy 
sets $A_i(i=1,2,\ldots)$ are formed and let $f_t$ be a collection of 
$\mu_{A_i}(t)(i=1,\ldots,m)$. Then, $f_t(t=0,1,\ldots,N)$ is called a 
fuzzy time series on $y_t(t=0,1,\ldots,N)$.

A fuzzy relationship between one point at time $(t)$ and its successor is 
represented by:

	\begin{equation}\label{eq:relationship}
	f_t \implies f_{t+1} \,.
	\end{equation}
	
We suggest a slightly more generic definition that can deal with segments. 
Indeed, we will consider fuzzy relationships between \textit{any element} at 
time $(t)$ and its successor, with $f_t$ being the segment $s^{(t)}$ and 
$f_{t+1}$ its segment $s^{(t+1)}$ (i.e., $f_t$ describes the \textit{entire} 
segment, instead of a specific point of the time series).

\subsection{Fuzzy inference systems}
\label{ssec:fis}
Fuzzy inference systems (FIS) can model uncertain and complex human reasoning 
tasks. FISs use ``IF antecedent THEN consequent'' rules as inference mechanism, 
where the antecedent and the consequent of the rule are linguistic terms that 
can handle multi-valued logic.

Different types of fuzzy inference systems exist. Two of them are widespread in 
the literature, namely the Takagi-Sugeno \cite{Sugeno1985} and the Mamdani 
\cite{Mamdani1974} type. The main difference between these two is that the 
latter uses output membership functions to describe linguistic terms, whereas 
the former uses output membership functions to describe crisp values. In this 
paper, the Mamdani inference system is considered because of its 
relative simplicity.

\begin{figure}
	\centering
	\includegraphics[width=\linewidth]{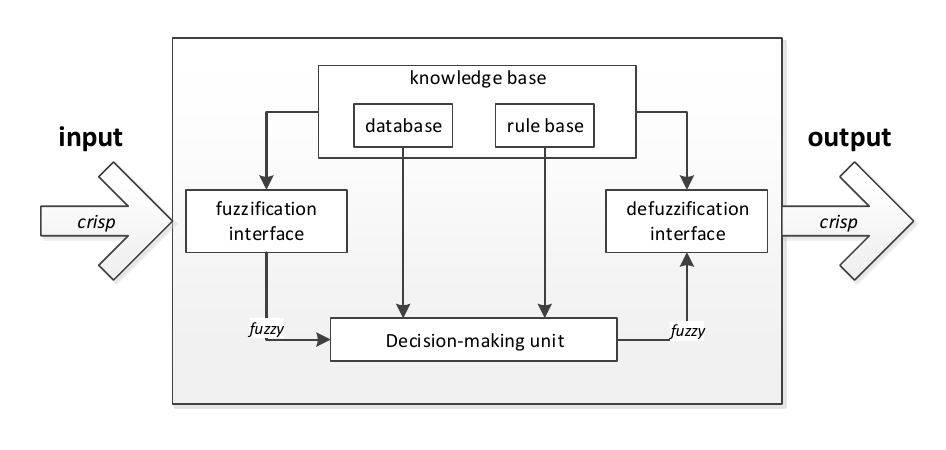}
	\caption{Structure of a fuzzy inference system. }
	\label{fig:fisStructure}
	\end{figure}

A fuzzy inference system is defined (see Fig. 
\ref{fig:fisStructure}) by a \textit{rule base} 
containing the set of ``IF-THEN'' rules; a \textit{database} with the fuzzy 
sets and their membership functions; a \textit{decision-making unit} performing 
inference  based on the rules; a \textit{fuzzification interface} transforming 
the crisp  inputs into degrees of match with linguistic values; and a 
 \textit{defuzzification interface} transforming the fuzzy results of the 
inference into numbers.

On top of this structure, the inference process is defined according to the 
5 following stages:

	\begin{enumerate}[1)]
	\item fuzzification of the inputs;
	\item combination of the antecedents with conjunction or disjunction 
	functions;
	\item rules firing and implication of 
	the consequent;
	\item aggregation of the consequent; and
	\item defuzzification of the output.  
	\end{enumerate}
	
\section[The proposed FCPD method]{The proposed method for the fuzzy detection 
of change points in crime-related time series\sectionmark{The proposed FCPD 
method}
}
\sectionmark{The proposed FCPD method}
\label{sec:proposed}

In this section, a novel method for the Fuzzy Change 
Points Detection in crime-related time series, FCPD, is proposed. Based on the 
particular shape space representation 
and the dynamic segmentation method described in Sect. 
\ref{sec:concepts}, a unique approach is proposed. 

FCPD consists of 2 steps:
	\begin{enumerate}[1)]
	\item the \textit{segmentation} of the time series by the means of the 
	shape space representation (represented in pseudo-code by Algorithm 
	\ref{alg:segmentation});
	\item the fuzzy \textit{querying}, by the use of linguistic variables, of 
	change points based on the discovered segments (represented in pseudo-code 
	by Algorithm \ref{alg:segmentation}).
	\end{enumerate}
	
These 2 steps are performed on-line. Algorithm 
\ref{alg:segmentation} starts with the very first observation of a time series 
and grows a window at every new observation. Every time a new segment is set 
(defined by some error criteria), Algorithm \ref{alg:querying} can be run to 
answer queries bases on the discovered segments (i.e., the outputs of Alg. 
\ref{alg:segmentation}). The FIS structure (i.e., the membership functions,
the linguistic variables and the rules) and the query are defined by the user 
in accordance with the use case and do not change over time.


	\begin{algorithm}[h]
	\DontPrintSemicolon
	\SetKwInOut{Input}{input}\SetKwInOut{Output}{output}
	\Input{$s=(y_0,y_1,\ldots,y_n)$, the time series\\
	$K$, the degree\\
	$\mathit{th}$, the thresholds}
	\Output{$\boldsymbol{\alpha}$, the coefficients}
	\Begin
	{
		$w \longleftarrow initWindow(s[0..K])$\;
		$c \longleftarrow initCoefs(w,K)$\;
		$i \longleftarrow K+1$\;
		
		\While{$i<=n$}
		{
			$w \longleftarrow growWindow(w,s[i])$\;
			
			$c \longleftarrow 
			updateCoefs(c,w,K)$\;
				
			\If{$newSegment(c,\mathit{th})$}
			{
				remove $s[0..i]$ from $s$\;
				add $c$ to $\boldsymbol{\alpha}$\;
				$\boldsymbol{\beta} \longleftarrow query(\boldsymbol{\alpha})$\;
				\textbf{goto \textit{begin}}\;
			}
			$i \longleftarrow i+1$\;
		}
	}
	\caption{\textsc{segmentation}, for on-line 
	segmenting\label{alg:segmentation}}
	\end{algorithm}

	\begin{algorithm}
	\DontPrintSemicolon
	\SetKwInOut{Input}{input}\SetKwInOut{Output}{output}
	\Input{$q$, the query (global variable)\\
	$F$, the FIS structure (global variable)\\
	$\boldsymbol{\alpha}$, the coefficients}
	\Output{$\boldsymbol{\beta}$, the sorted segments}
	\Begin
	{	
		$FIS \longleftarrow initFIS(F)$\;
		$scores \longleftarrow inferFIS(FIS,\boldsymbol{\alpha},q)$ 'see 
		Fig. \ref{fig:fisStructure} \;
		$\boldsymbol{\beta} \longleftarrow sortSegments(scores)$\;
		
	}
	\caption{\textsc{query}, for on-line change points 	
	querying\label{alg:querying}}
	\end{algorithm}

\subsection*{\textbf{Step 1: Finding the segments $W(s)$ of the time series}}
Starting with a time series represented by $s=(y_0,y_1,\ldots,y_N)$ as in Eq. 
\ref{eq:s}, the parameter $K$ (i.e., the degree of 
the polynomials), and the thresholds, the segmentation process (Sect. 
\ref{ssec:seg}) is iteratively applied. First, a growing window is positioned 
on the first element ($y_0$), the polynomial expansion is computed and its 
coefficients $\boldsymbol{\alpha}$ are extracted according to the chosen degree 
$K$. Based on these coefficients some segmentation criteria can be defined. Two 
of them are hereafter suggested.

The first threshold is the deviation of the predicted 
value:

	\begin{equation}
	c_{\mathit{DPU}}=\left\{
		\begin{array}{l l}
	    1, & \quad \text{if $|p(x_t)-y_t| > th_{\mathit{DPU}}$},\\
	    0, & \quad \text{otherwise},
		\end{array} \right.
	\end{equation}
	
\noindent where $p(x_t)$ is the predicted value of the regression, $y(x_t)$ 
the actual value at time $t$ (the last point of the growing window), and 
$\mathit{th_{DPU}}$ the value of the threshold. The second is the
counting of the sign switches of the slope:

	\begin{equation}
	c_{\mathit{SSS}}=\left\{
	  \begin{array}{l l}
	  1, & \quad \text{if $\mathit{SSS} > th_{\mathit{SSS}}$},\\
	  0, & \quad \text{otherwise},
	  \end{array} \right.
	\end{equation}
	
\noindent where $\mathit{SSS}$ counts the number of sign switches of the 
slope within the window, that is, $\mathit{SSS}$ is incremented if a change in 
the sign of the slope is observed, and $\mathit{th_{SSS}}$ the value of the 
threshold. 
A new segment $s^{(i)}$ is added to $W(s)$ if these criteria are met (one 
single criterion can be enough), and the 
segment is then represented by the last $\boldsymbol{\alpha}$ computed, 
$\alpha^{(t)}$. Otherwise, the window is grown by adding the next point and the 
same steps are repeated until the end of the time series (i.e., 
$\boldsymbol{\alpha}$ is updated and the new criteria are again compared to 
the same thresholds). The result of this step is the set of continuous windows 
$W(s)$, their respective coefficients $\boldsymbol{\alpha}^{(i)}$ and their 
change points $\hat{x}^{(i)}$.

%

\subsection*{\textbf{Step 2: Querying the existence of particular change points 
with a fuzzy inference system}}
A query consists of the expression of geometric properties with linguistic 
values related to the coefficients, and the result of 
the query is the corresponding ``relevance score'' of each segment regarding 
the query. The query is specified with a fuzzy ``IF-THEN'' rule. To 
evaluate the query, features related to the coefficients from the subspace 
representation are given as input to the FIS, the query is added to the rule 
base, and the system infers the output. The defuzzification of the output is 
the answer to the original query. The membership functions of the FIS have to 
be specified at the beginning. In fact, linguistic variables and membership 
functions are part of the query, specified by the user to detect change points 
with regard to their applications. These parameters should not change over 
time, unless if the query itself changes. Setting the FIS amounts to:

	\begin{enumerate}[a)]
	\item \textit{Choosing the input(s) and the output(s) of the fuzzy 
	inference system in relevance of the query.} \\
	Inputs constitute the antecedent part of the rules and outputs the 
	consequent. As initial intuition, inputs could be the 
	coefficients $\boldsymbol{\alpha}^{(i)}$. The use of these coefficients 
	enables the expression of linguistic terms concerning the average, the 
	slope, or the curvature of the segments in the antecedent part of the rules.
	
	However, to handle queries considering more aspects, other input variables 
	can be 	considered to get different expressions. Relations 	between two	
	\textit{elements} (as in Eq. \ref{eq:relationship}, where $f_t 
	\implies f_{t+1}$), can also be inputs. For instance, if the element is a 
	segment, the input of the FIS is then the coefficient variation between two 
	segments, defined as:
	
		\begin{equation}
		v_{\alpha_k}(s^{(t)},s^{(t+d)})=\frac{\alpha_k^{(t+d)}-
		\alpha_k^{(t)}}{\alpha_k^{(t)}} \,,
		\end{equation}
		
	\noindent where $s^{(i)}$ is the segment of index $i$ in $W(s)$, 
	$\alpha_k^{(i)}$ the coefficient of order $k$ of this segment, and 
	$d$ the delay operator of segments, typically set to $1$. For the sake 
	of simplicity, these variations will be referred to as
	
		\begin{equation} \label{eq:variation}
		v_{\alpha_k}^{i->i+d} \,. 
		\end{equation}
		
	\noindent In other words, the use of variations, instead of the 
	coefficients 
	input to the FIS, enable to express relative changes between 
	two periods instead of absolute changes of value only.
	
	Other inputs to consider are for instance the size of each segment, the 
	variation of the size, or a set of primitive shapes. A combination of these 
	is also possible.
	
	The output of the FIS is more straightforward. Given the rules, the FIS 
	outputs the degree of similarity of each input to the geometric properties 
	specified in the query.	Therefore, only one output ---the relevance of the 
	query--- is assumed to be necessary in most cases. A FIS with several 
	outputs is nonetheless possible.
	

	
	\item \textit{Defining the linguistic terms and their membership 
	functions.} 
	\\	
	Each input/output of the inference system is generally defined by 
	multiple fuzzy sets. For example, $(\mathit{LOW})$, $(\mathit{MEDIUM})$, 
	and $(\mathit{HIGH})$ can be fuzzy sets for the coefficients as input, 
	whereas $(\mathit{DECREASE})$, $(\mathit{CONSTANT})$, and 
	$(\mathit{INCREASE})$ are sets of variations between elements. 
	For these terms we need membership functions that can be valued, as part of 
	the fuzzification and defuzzification process.
	
	\item \textit{Defining the inference rules.} \\
	The rules added to the inference system are the heuristics that guide the 
	search to find the appropriate change points. These heuristics use the 
	geometric estimators from the coefficient to express visual criteria of the 
	researched segments. These inputs are evaluated in the antecedent of the 
	rule and the output in the consequent. A weight can be added to each rule, 
	giving different degrees of importance in accordance with the confidence of 
	the heuristic.
	
	
	\item \textit{Inferring the output(s).} \\
	Infer the output(s) of the FIS (as described in Sect. \ref{ssec:fis}). 
	According to the rules, the segments which are the most relevant to the 
	query output a higher membership of the consequent.
	

	\end{enumerate}


	

\section{Empirical Evaluation} 
\label{sec:experiments}
This section provides an empirical evaluation of the proposed method, with a  
focus on crime data. For the sake of an overall evaluation, different types of 
time series with different objectives are analyzed. 

First, qualitative analyses, representing illustrated case studies helping 
practitioners to better evaluate the method, are conducted 
(with a total of 4 time series):
	\begin{enumerate}[1)]
	\item the analysis of cyclic data, to illustrate the use of the proposed 
	method in a simple environment;
	\item a case study of crime trends monitoring, to support the validity and 
	applicability of flexible change points detection and querying according to 
	the domain of crime analysis;
	\item the analysis of the \textit{TOPIX} time series (financial real-world 
	data), in a financial case study, to test the domain-free applicability of 
	FCPD;
	\end{enumerate}

Second, quantitative analyses, each time systematically compared with two 
comprehensive data sets (with a total of 96 time series):
	\begin{enumerate}[1)]
	\setcounter{enumi}{3}
	\item a comparison with a similar change points detection algorithm, BFAST, 
	is performed on both the CICOP and the SWX data sets for assessing the 
	accuracy and the complexity of FCPD;
	\item a sensitivity analysis on both the CICOP and SWX data sets is carried 
	on for measuring the impact of the parameters on the results of the 
	proposed method. 
	\end{enumerate}
	
These two data sets used in the quantitative part are the following:
	\begin{enumerate}[A)]
	\item the \textit{CICOP} data set (crime real-world 
	data), consisting of 32 time series each with 70 observations. These 
	time series describe monthly events for a period of 6 years (2009-2014) 
	of serial- and itinerant-related crimes (such as burglaries);
	\item the \textit{SWX} data set (financial real-world data), consisting 
	of 64 time series each with 120 observations. These time series describe
	monthly stock data for a period of 10 years (2005-2014) from the small 
	and medium capitalizations of the SWX (Swiss Exchange Market).
	\end{enumerate}


For these experiments, a MATLAB version of FCPD has been implemented by 
the authors. Time series were normalized ($\mu=0$, $\sigma^2=1$) before 
analysis. The only reason for normalizing time series is to make comparison 
between different data sets easier: indeed, normalizing time 
observations leads to consistent thresholds throughout all time series (i.e., 
thresholds in accordance with the mean and the variance of the time series).

\subsection{Simulation with cyclic data}
This first experiment aims to detect simple change points within the cycle time 
series. For that purpose, we generated a normalized sinusoidal time series of 
$2000$ data points, representing a cyclical activity. We introduced two 
``anomalies'', the first 
between the x-interval $[500, 600]$ by adding noise with a standard normal 
distribution ($N(0,1)$) to the observed values and the second between 
$[1400,1600]$ by replacing the observed values with $y=0.5+u/2$, where $u$ is a 
noise factor with a standard normal distribution.

	\begin{figure}[htp]
	\centering
	\includegraphics[width=\linewidth]{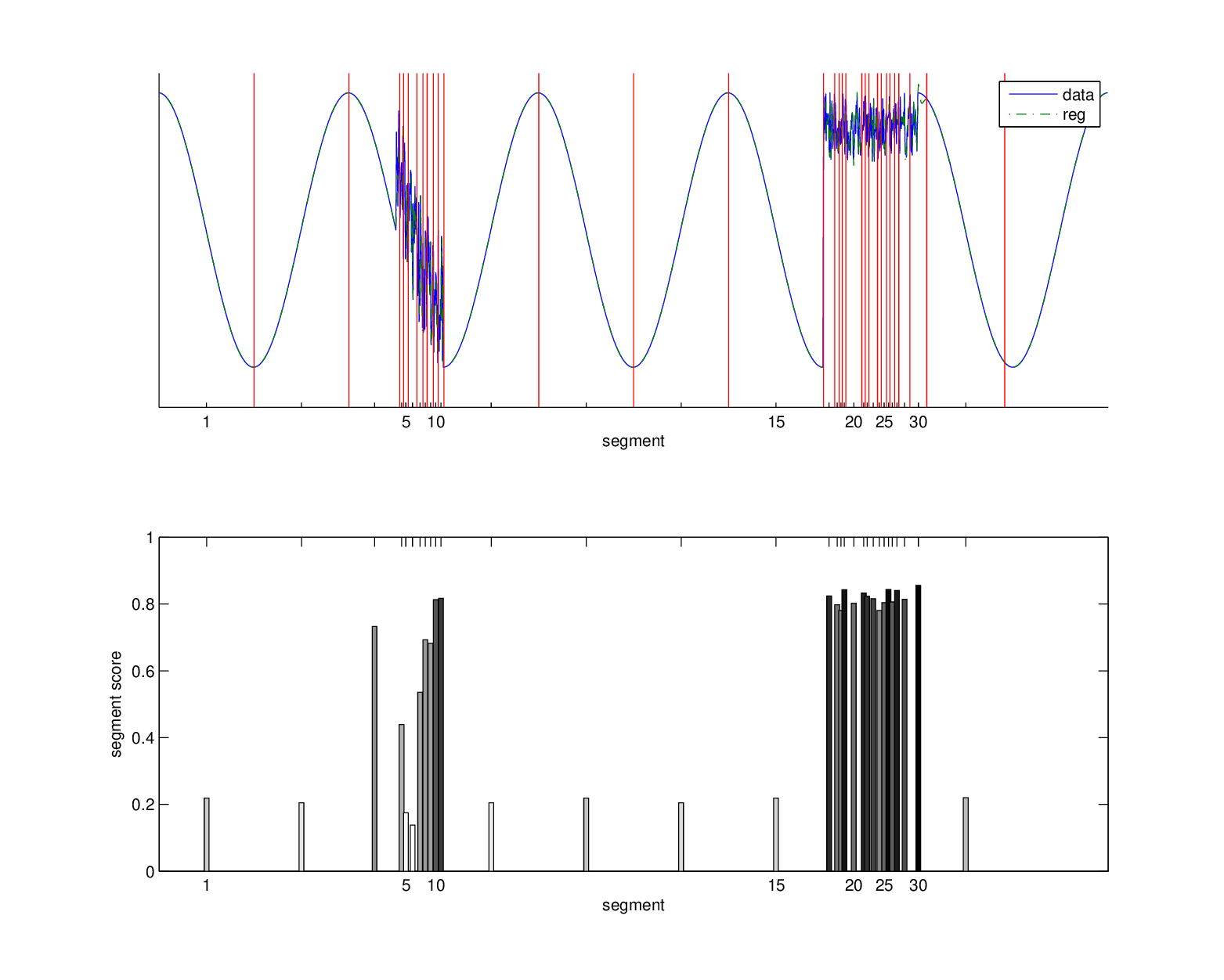}
	\caption[The cycle time series.]{The cycle time series. \textit{(Top)} 
	Segmentation 
	of the time series, $31$ change points detected. \textit{(Bottom)} 
	Output of the FIS for each segment, representing the relevance of the 
	query.}
	\label{fig:fisCycleBreak}
	\end{figure}

\begin{figure}[htp]
	\centering
	\includegraphics[width=\linewidth]{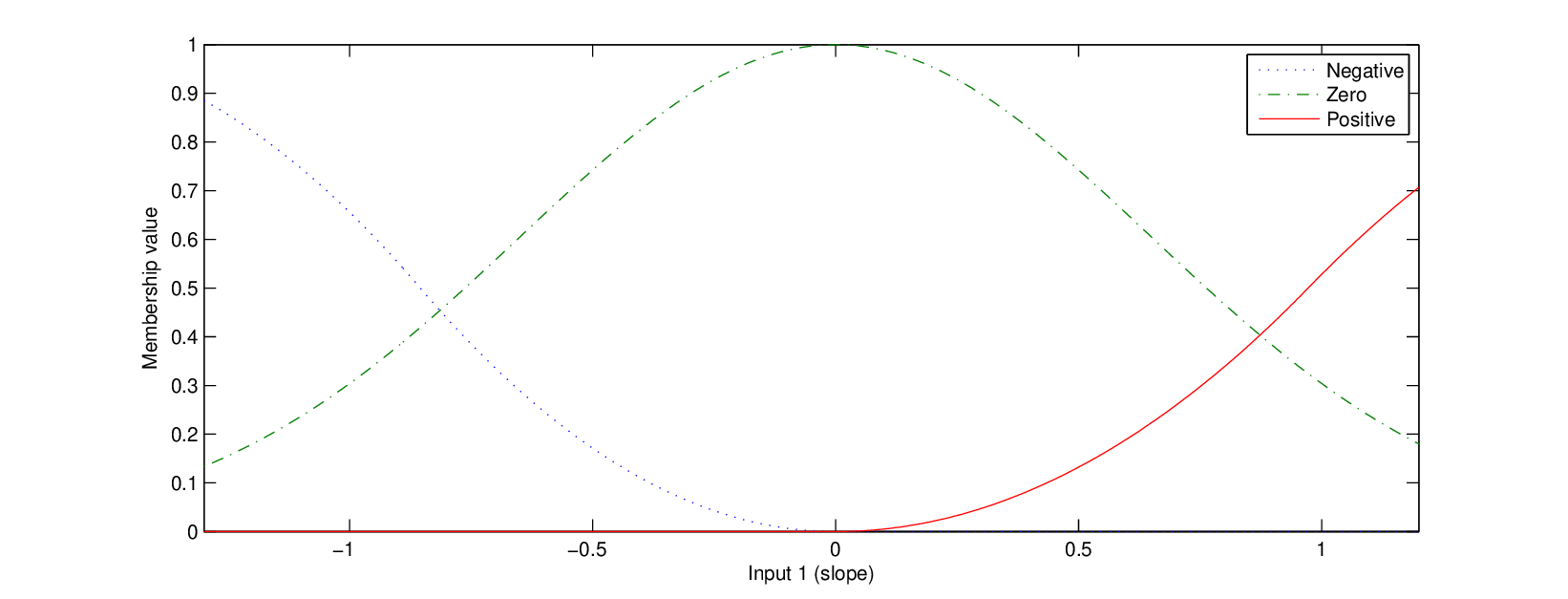}
	\caption{Input membership functions of the cycle time series.}
	\label{fig:mfCycleBreak_a}
	\end{figure}

\begin{figure}[htp]
	\centering
	\includegraphics[width=\linewidth]{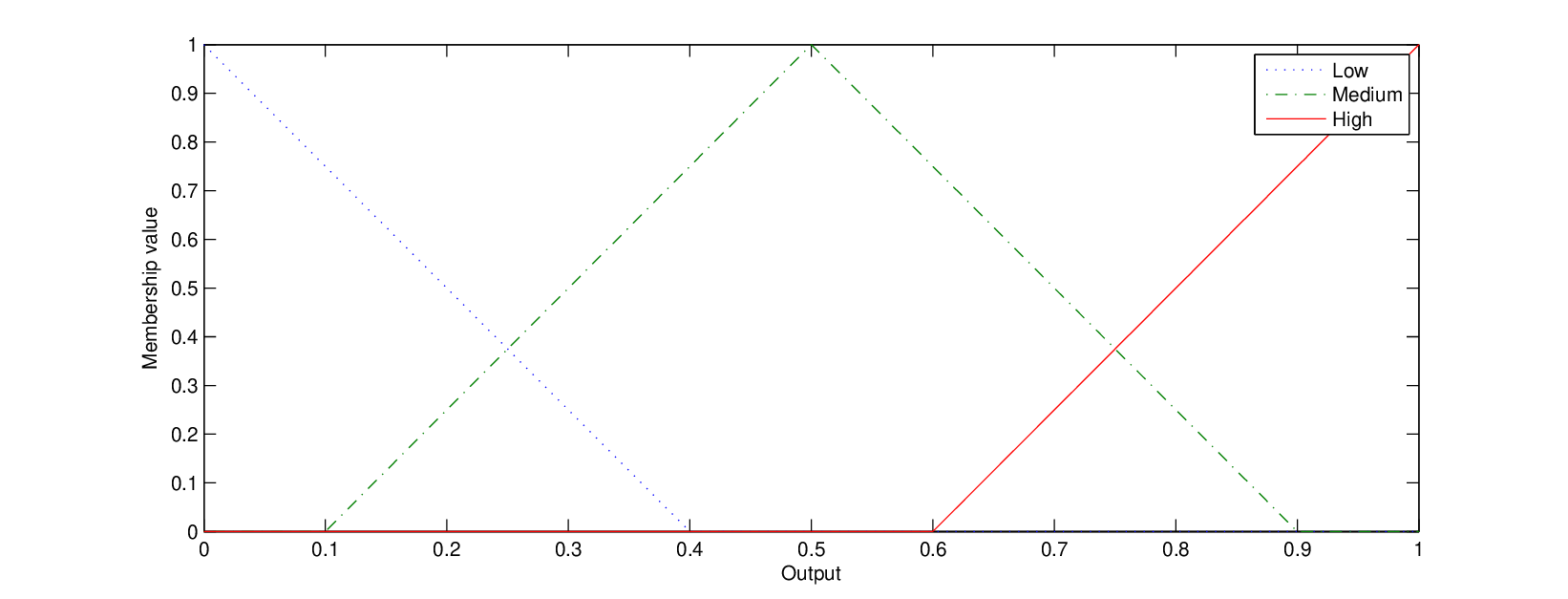}
	\caption{Output membership functions of the cycle time series.}
	\label{fig:mfCycleBreak_b}
	\end{figure}

For the segmentation step, we set the maximum degree 
of the polynomial regression $K$ equals to $5$ and a single threshold 
for the switches of the sign slope ($\mathit{th_{SSS}}$) to the value of $1$. 
As a result, $31$ 
change points were detected (top of Fig. \ref{fig:fisCycleBreak}). We then 
used as input to the FIS the coefficient matrix for the 
average (i.e., the coefficients of degree $0$ for each segment). The input is 
described with $3$ different linguistic terms, namely $\mathit{negative}$, 
$\mathit{zero}$, and $\mathit{positive}$, respectively represented by a 
Z-shaped, a Gaussian, and a S-shaped membership function as depicted in 
Fig. \ref{fig:mfCycleBreak_a}. The output membership is a triangular-shaped 
function, using $\mathit{low}$, $\mathit{medium}$, and 
$\mathit{high}$ as fuzzy sets to denote the relevance of the queried geometric 
properties (Fig. \ref{fig:mfCycleBreak_b}). For the inference part, we chose 
the $\mathit{min}$ function 
for the implication, the $\mathit{max}$ function for the aggregation, and the 
$\mathit{centroid}$ function for the defuzzification. The query for identifying 
change points in the cycle is then modeled through the following rules:

\begin{enumerate}[a)]
\item IF (\textit{average} is not \textit{zero}), THEN (\textit{score} is 
\textit{high})
\item  IF (\textit{average} is \textit{zero}), THEN (\textit{score} is 
\textit{low}).
\end{enumerate}

These two basic rules use the average of each segment to determine the 
degree of change within the cycle. The output of the FIS (bottom of Fig. 
\ref{fig:fisCycleBreak}) describes a score within the $[0,1]$ interval for each 
segment of the time series. High scores are produced for the values where some 
noise was added (segments $3$ to $11$, and segments $16$ to $30$), which 
confirms the expected results.
	
\subsection{Case study: crime trends monitoring}
\label{ssec:ctm}
The objective of crime trends monitoring is to automatically detect change 
points within the development of crime. This case study shows how crime 
analysts can monitor sudden changes in the number of crimes, allowing them 
better to allocate resources (e.g., by sending dedicated patrols when a rise is 
detected).  We want to emphasize that the proposed method is used on-line, 
meaning that we do not need the entire time series to perform these analyses 
and the results only depend on past values.

\begin{figure}[htp]
	\centering
	\includegraphics[width=\linewidth]{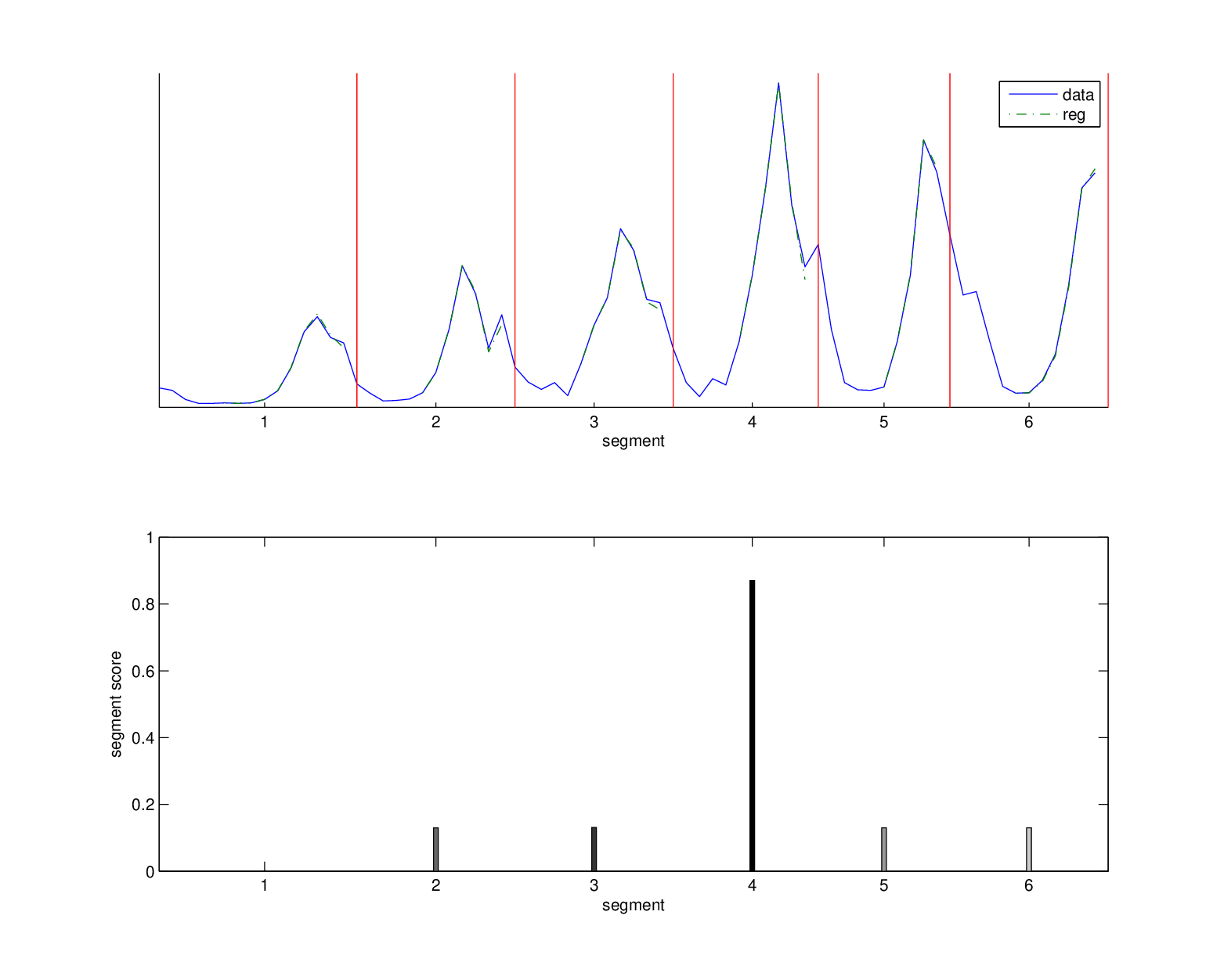}
	\caption[Evening burglaries from the CICOP data set.]{Evening burglaries 
	time series from the CICOP data set. 
	\textit{(Top)} 
	Segmentation of the time series, with $6$ change points detected. 
	\textit{(Bottom)} Output of the FIS for each segment, representing the 
	relevance of the query (i.e., changes in trend).}
	\label{fig:fisCicopSera}
	\end{figure}

\begin{figure}[htp]
	\centering 
	\includegraphics[width=\linewidth]{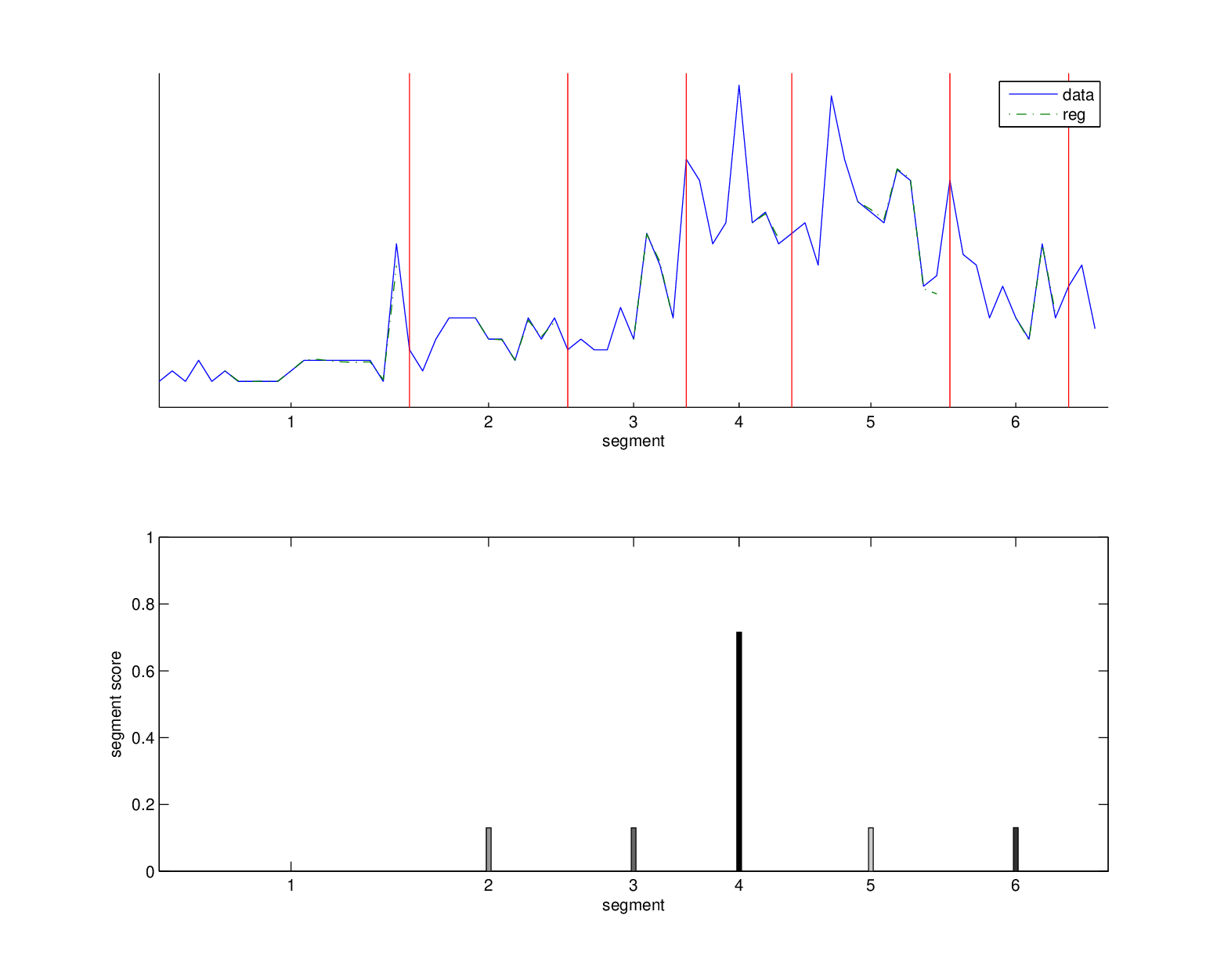}
	\caption[ATM break-ins time series from the CICOP data set.]{ATM break-ins 
	time series from the CICOP data set. 
	\textit{(Top)} 
	Segmentation of the time series, with $6$ change points detected. 
	\textit{(Bottom)} Output of the FIS for each segment, representing the 
	relevance of the query (i.e., changes in trend).}
	\label{fig:fisCicopBancomat}
	\end{figure}

\begin{figure}[htp]
	\centering
	\includegraphics[width=\linewidth]{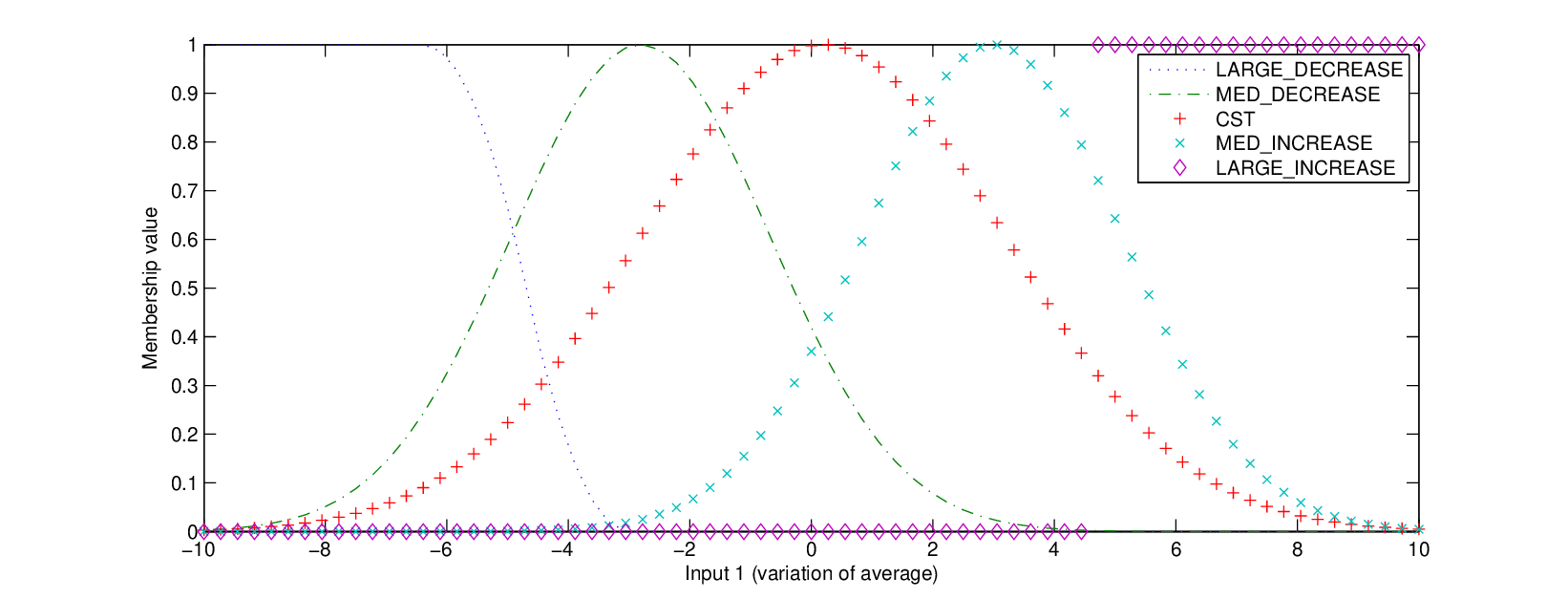}
	\caption{Input membership functions of both evening burglaries and ATM 
	burglaries time series.}
	\label{fig:mfCicopSera_a}
	\end{figure}

For this purpose, we illustrate the detection of change points with two time 
series from the CICOP data set. The first time series describes 
evening burglaries of individual houses or flats, with 72 monthly data points 
for a period of 6 years (top of Fig. \ref{fig:fisCicopSera}). The second time 
series represents ATM break-ins, with the same sampling (top of Fig. 
\ref{fig:fisCicopBancomat}). A more detailed description of this data set can 
be found in \cite{Albertetti2013b}. The segmentation settings are identical for 
both time series of the data set: values are normalized, two disjunctive 
thresholds are set ($\mathit{th_{DPU}}=0.05$ and $\mathit{th_{SSS}}=2$; only 
one threshold need to be exceeded to set a new segment), and 
$K$ is set to $5$. The input of the FIS is the coefficient \textit{variation} 
between two consecutive segments ($v_{\alpha_k}^{i->i+1}$, as in Eq. 
\ref{eq:variation}) of the average coefficient, with $5$ fuzzy sets (the 
membership functions are
shown in Fig. \ref{fig:mfCicopSera_a}).

The same three inference rules for both time series were given as heuristics to 
querying changes in the trend:

	\begin{enumerate}[a)]
	\item IF (\textit{var\_average} is \textit{large\_decrease}), THEN 
	(\textit{score} is	\textit{high})
	\item  IF (\textit{var\_average} is \textit{large\_increase}), THEN 
	(\textit{score} is	\textit{high})
	\item  IF (\textit{var\_average} is \textit{constant}), THEN 
	(\textit{score} is	\textit{low}).
	\end{enumerate}

The highest score for both time series is observed in segment $4$. Indeed, both 
time series are suggesting a high variation of the average after segment $3$; 
in the next segments, the trends remain pretty stable and the score remains low.


\subsection{Case study: change points detection on the financial TOPIX time 
series}
\label{ssec:topix}
To assess the general applicability of the method in a 
different domain, we used the TOPIX time series to detect and query change 
points. The weekly values consist of 522 data points 
(years 1985 to 1994) from the TOPIX index (\textit{TPX:IND}, i.e., the Tokyo 
Stock Exchange Price Index). We also compare our results with the work of 
Yamanishi and Takeuchi in \cite{Yamanishi2002} and \cite{Takeuchi2006}, which 
have used the same time series for change points and outliers detection. 

The purpose of this case study is to detect change points considered as the 
steepest slopes of the considered time-frame. In the financial domain, it 
proves very useful to find change points in real-time, giving the investors a 
trend indicating the right moment for selling or buying stocks.

	\begin{figure}[htp]
	\centering
	\includegraphics[width=\linewidth]{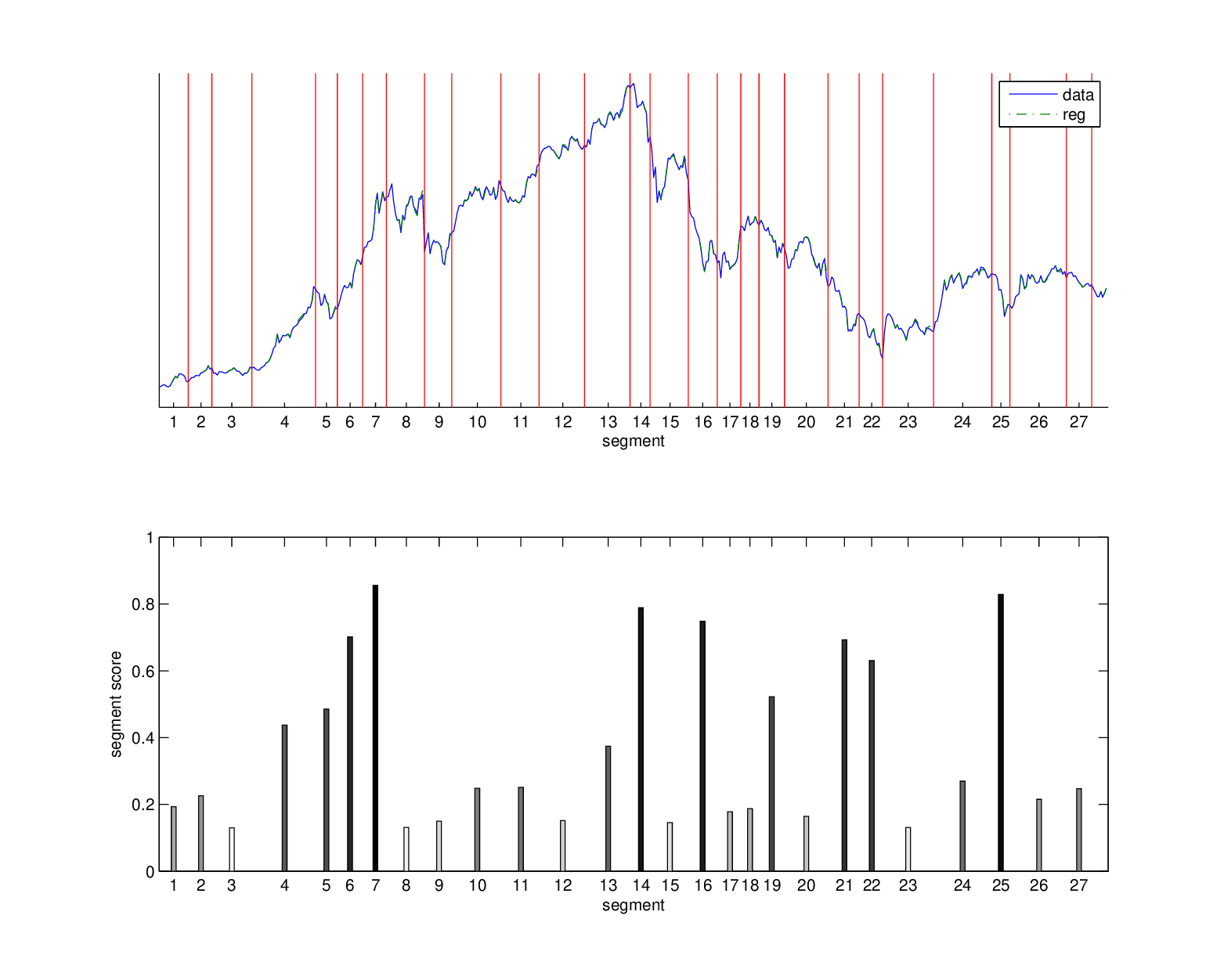}
	\caption[The TOPIX time series.]{The TOPIX time series. \textit{(Top)} 
	Segmentation of the time 
	series, 
	with $27$ change points detected. \textit{(Bottom)} Output of the FIS for 
	each segment, representing the relevance of the query (i.e., finding the 
	steepest slopes).}
	\label{fig:fisTopix}
	\end{figure}

For the segmentation process, we set $K$ to $5$ and 
two independent thresholds ($\mathit{th_{SSS}}=2$ and $\mathit{th_{DPU}}=0.05$ 
respectively for the switches of the sign slope and the deviation of the 
predicted value). Figure 
\ref{fig:fisTopix} (top) shows the $27$ change points detected.

\begin{figure}[htp]
	\centering
	\includegraphics[width=\linewidth]{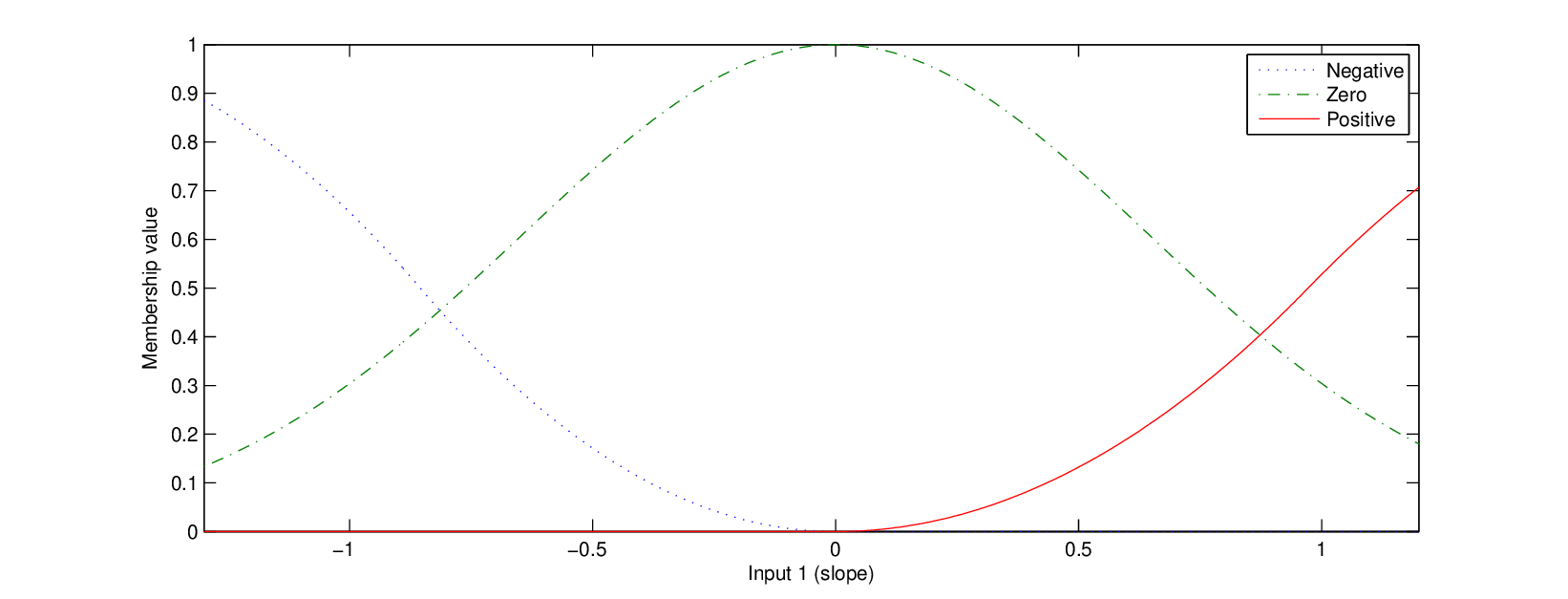}
	\caption{Input membership functions of the TOPIX time series.}
	\label{fig:mfTopix_a}
	\end{figure}

The querying step is used for identifying steep slopes. As input 
to the FIS, the slope coefficient is used. The fuzzy sets describing the input 
are $\mathit{negative}$, $\mathit{zero}$, and $\mathit{positive}$ (Fig. 
\ref{fig:mfTopix_a}). The output 
membership is the same as for the cycle time series (Fig. 
\ref{fig:mfCycleBreak_b}). Three inference rules were given as heuristics to 
describe a steep slope shape:

	\begin{enumerate}[a)]
	\item IF (\textit{slope} is \textit{negative}), THEN (\textit{score} is 
	\textit{high})
	\item  IF (\textit{slope} is \textit{positive}), THEN (\textit{score} is 
	\textit{high})
	\item  IF (\textit{slope} is \textit{zero}), THEN (\textit{score} is 
	\textit{low}).
	\end{enumerate}
	
Segments $\#7$, $\#25$, $\#14$, and $\#16$ have the highest score with FCPD 
(see Table \ref{tab:significantPoints}). We want to emphasize that these 
results depend on the segmentation method, as the slope of the segment is the 
average of the slopes of the data points belonging to the segment.

	\begin{table}[htp]
	\caption[Top 4 segments in the TOPIX time series.]{Top 4 segments in the 
	TOPIX time series identified as the 
	\textit{significant changes} with the proposed method. Segment intervals 
	are specified by values on the x-axis.}
	\label{tab:significantPoints}
	\centering
		\begin{tabular}{cccc}
		\toprule[0.1em]
		Rank & Number & Interval & Score 
		\\ \toprule 
		1 & 7 & [112,125] & 0.856 
		\\ 
		2 & 25 & [458,468] & 0.829 
		\\ 
		3 & 14 & [259,270] & 0.789
		\\ 
		4 & 16 & [291,307] & 0.748 
		\\ \bottomrule[0.1em] 
		\end{tabular} 
	\end{table}

In their experiment \cite{Yamanishi2002}, Yamanishi and Takeuchi highlight 
$4$ significant changes, occurring in our resulting segments $\#8$, $\#2$, 
$\#14$, and $\#25$. These results are very similar to ours (i.e., segment $\#7$ 
from FCPD occurs in segment $\#8$ with Yamanishi et al., segment $\#24$ in 
segment $\#25$, segment $\#14$ is identical, and segment $\#16$ has no direct 
correspondence).

\begin{figure}[htp]
	\centering
	\includegraphics[width=0.9\linewidth]{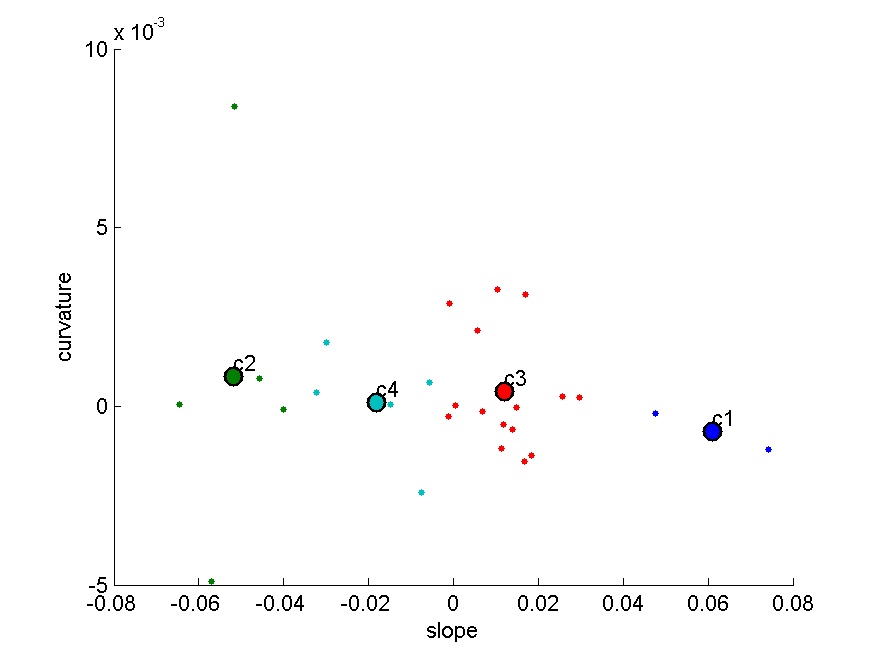}
	\caption[Clustering of the segments from the TOPIX time series.]{Clustering 
	of the segments from the TOPIX time series. The 
	segments are described with their slope and their curvature.}
	\label{fig:clusteringTopix}
	\end{figure}

\begin{figure}[htp]
	\centering
	\includegraphics[width=\linewidth]{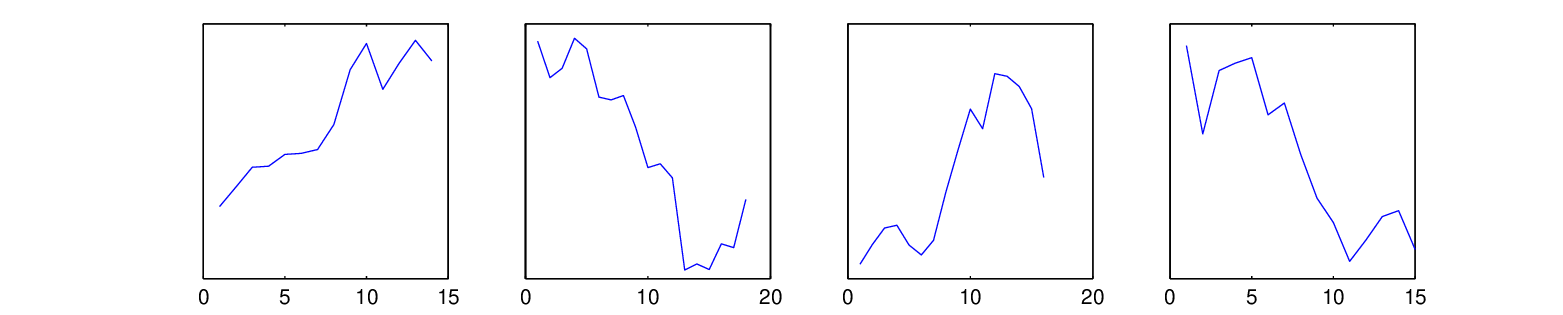}
	\caption{Primitive shapes of the TOPIX time series, resulting from the 
	clusters}
	\label{fig:primitivesTopix}
	\end{figure}

Besides, we also want to illustrate that FCPD is not limited to change 
points detection. Indeed, the shape space representation can be used to perform 
other types of analysis based on the meaningful distance computed with the 
shape space representation, such as clustering. In our example, we attempt to 
discover basic/primitive shapes in the time series. For that, we apply the 
K-means algorithm to the slope and curvature coefficients of the segments (i.e. 
the output of step 1 of the proposed method, Sect.\ref{sec:proposed}), with the 
number of clusters set to $4$ (with the objective to delineate both negative 
and positive clusters of slopes and curvatures). The centroids are depicted in 
Fig. \ref{fig:clusteringTopix}. The 
closest segments to the centroids, identified as the $4$ potential primitive 
shapes, are shown in Fig. \ref{fig:primitivesTopix}. One should notice that in 
this case the slope variable contains more information on the shape than the 
curvature variable.

\subsection{Comparison with the BFAST algorithm}
\label{ssec:comparison}
Our proposed method is compared to the ``break for additive seasonal and 
trend'' (BFAST) algorithm (\cite{Verbesselt2010} and \cite{Verbesselt2010b}) in 
terms of similarity of the three most important change points detected by 
each method. BFAST has originally been developed for detecting changes in 
phenology, more precisely for climatic variations from remote sensor data. The 
method is however not specific to a particular data type. It 
uses the seasonal-trend decomposition procedure based on Loess (STL) and an 
estimation of breaks based on \cite{Bai1994} with least-squares. It basically 
performs two steps, each being off-line, requiring access to past and future 
values in the computation window: first, the seasonal component is computed and 
removed from the observed data and second, the breakpoints are estimated.

We used the implementation of the R package \textit{bfast} with standard 
parameters ($\mathit{h=0.15}$, $\mathit{max.iter=1}$, 
$\mathit{season="harmonic"}$, 
$\mathit{breaks=3}$) 
to find a maximum of $3$ significant change points in both the CICOP and the 
SWX data sets.

\begin{figure}[htp]
	\centering
	\includegraphics[width=\linewidth]{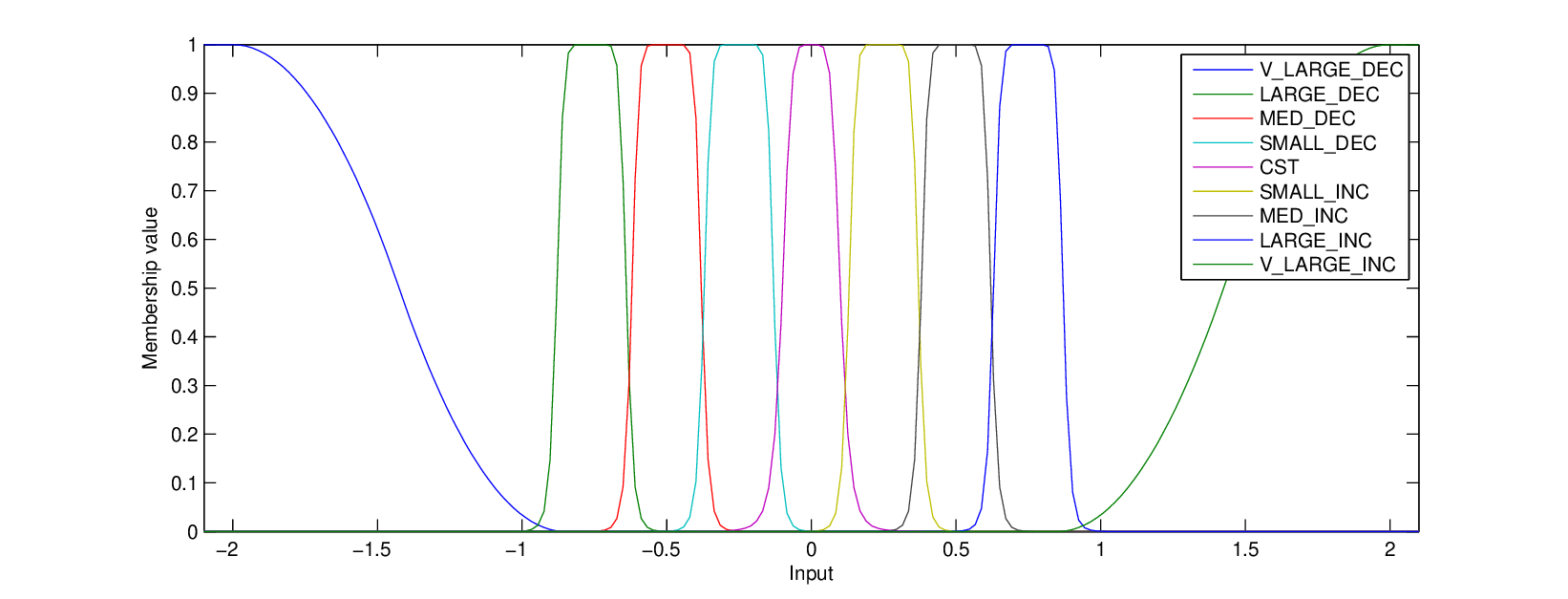}
	\caption{Input membership functions for the FIS of the CICOP and SWX data 
	set.}
	\label{fig:mfStandard_input}
	\end{figure}

\begin{figure}[htp]
	\centering
	\includegraphics[width=\linewidth]{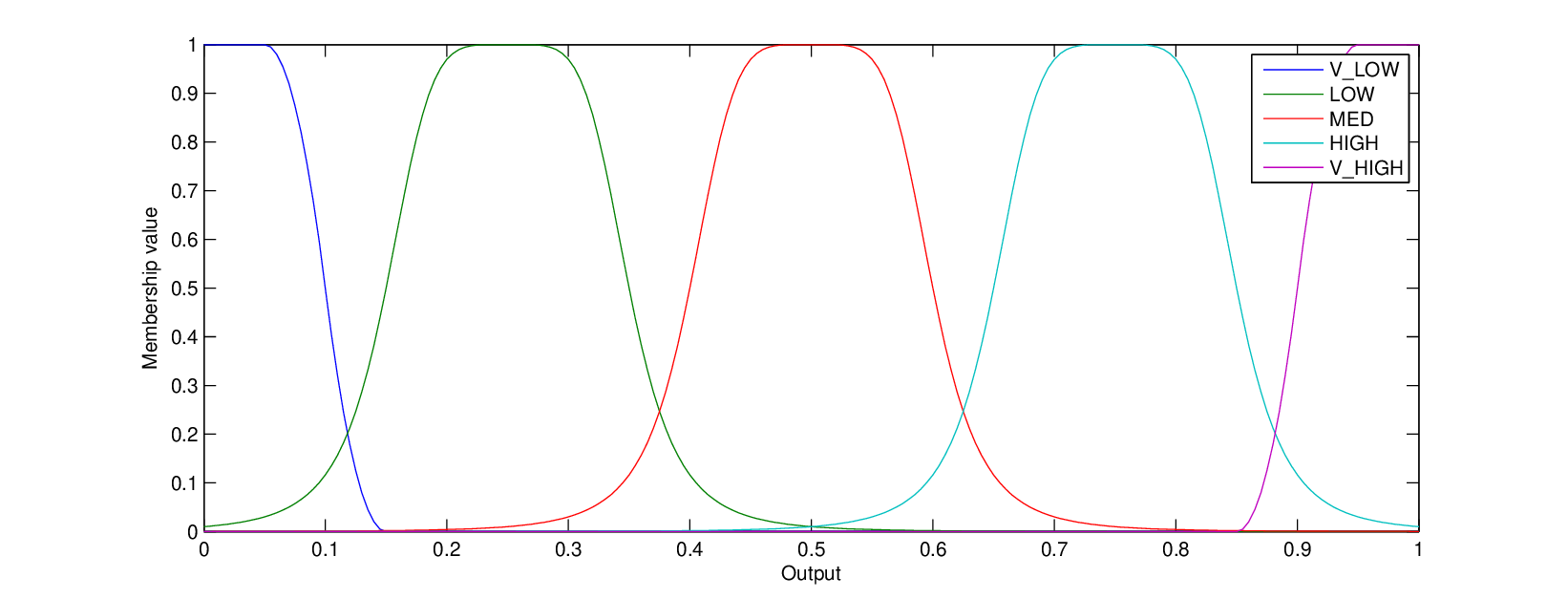}
	\caption{Output membership functions for the FIS of the CICOP and SWX data 
	set.}
	\label{fig:mfStandard_output}
	\end{figure}

The settings in FCPD are the same for both data sets, as the nature of these 
two data sets are similar and the objectives are identical. We set a threshold 
for the deviation of the predicted value ($th_{DPU}$) of $0.11$ and $K$ to 5.
Fig. \ref{fig:mfStandard_input} depicts the uniform membership functions for 
the input variables and Fig. \ref{fig:mfStandard_output} for the output 
variable. The rules are simply relating the degree of variation to the degree 
of change, considering both the average and the slope:

	\begin{enumerate}[a)]
	\item IF (\textit{var\_average} or \textit{var\_slope} is 
	\textit{very\_large\_decrease}),
	THEN (\textit{score} is	\textit{very\_high})
	\item IF (\textit{var\_average} or \textit{var\_slope} is 
	\textit{large\_decrease}),
	THEN (\textit{score} is	\textit{high})
	\item IF (\textit{var\_average} or \textit{var\_slope} is 
	\textit{medium\_decrease}),
	THEN (\textit{score} is	\textit{medium})
	\item IF (\textit{var\_average} or \textit{var\_slope} is 
	\textit{small\_decrease}),
	THEN (\textit{score} is	\textit{low})
	\item IF (\textit{var\_average} or \textit{var\_slope} is 
	\textit{constant}),
	THEN (\textit{score} is	\textit{very\_low})
	\item IF (\textit{var\_average} or \textit{var\_slope} is 
	\textit{small\_increase}),
	THEN (\textit{score} is	\textit{low})
	\item IF (\textit{var\_average} or \textit{var\_slope} is 
	\textit{medium\_increase}),
	THEN (\textit{score} is	\textit{medium})
	\item IF (\textit{var\_average} or \textit{var\_slope} is 
	\textit{large\_increase}),
	THEN (\textit{score} is	\textit{high})
	\item IF (\textit{var\_average} or \textit{var\_slope} is 
	\textit{very\_large\_increase}),
	THEN (\textit{score} is	\textit{very\_high})
	\end{enumerate}
	
\begin{figure*}[htp]
		\captionsetup[subfigure]{position=b}
		\centering
		\caption{Illustration of the comparison of the 3 most significant 
			changes between the BFAST algorithm and the proposed method.}
		\label{fig:comparison_illustration}
		\subcaptionbox{The 3 most important change points detected by BFAST.}
		{\includegraphics[width=.33\linewidth]{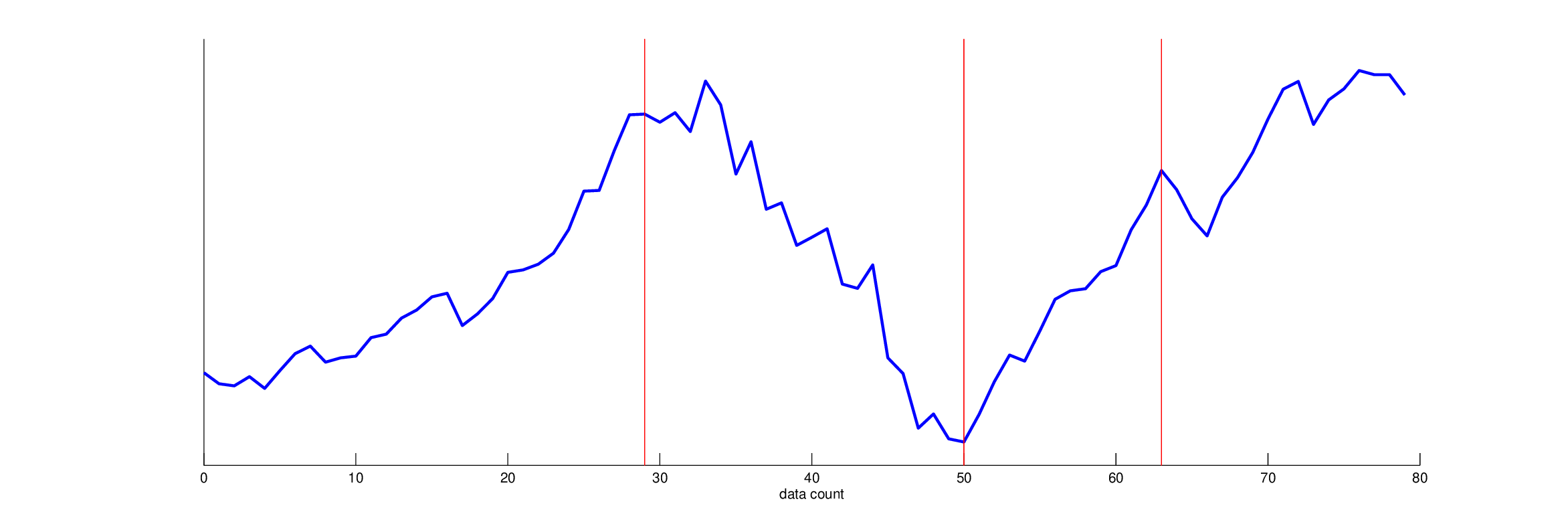}}
		\hfill
		\subcaptionbox{The 3 most important change points detected by FCPD.}
		{\includegraphics[width=.33\linewidth]{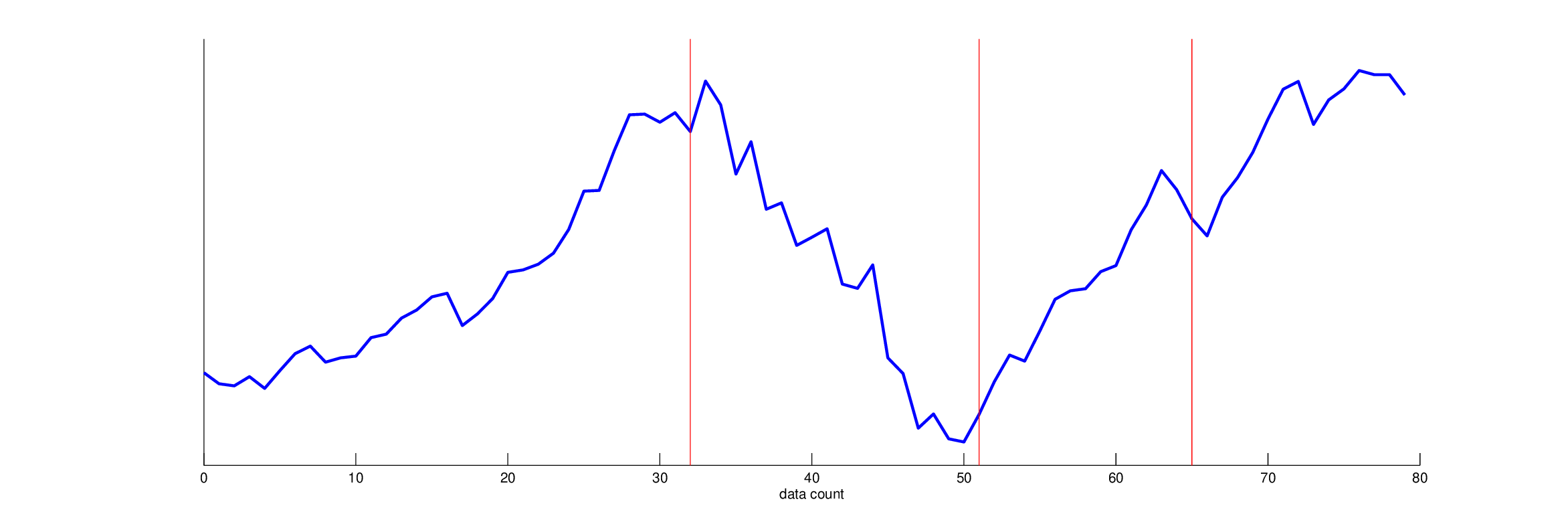}}
		\hfill
		\subcaptionbox{Calculation of the occurrence offset (red areas, from 
		left 
			to right): the 1st offset is reported as $+3$ data points in the 
			x-axis 
			for FCPD, the 2nd as $+1$, and the 3rd as $+2$.}
		{\includegraphics[width=.33\linewidth]{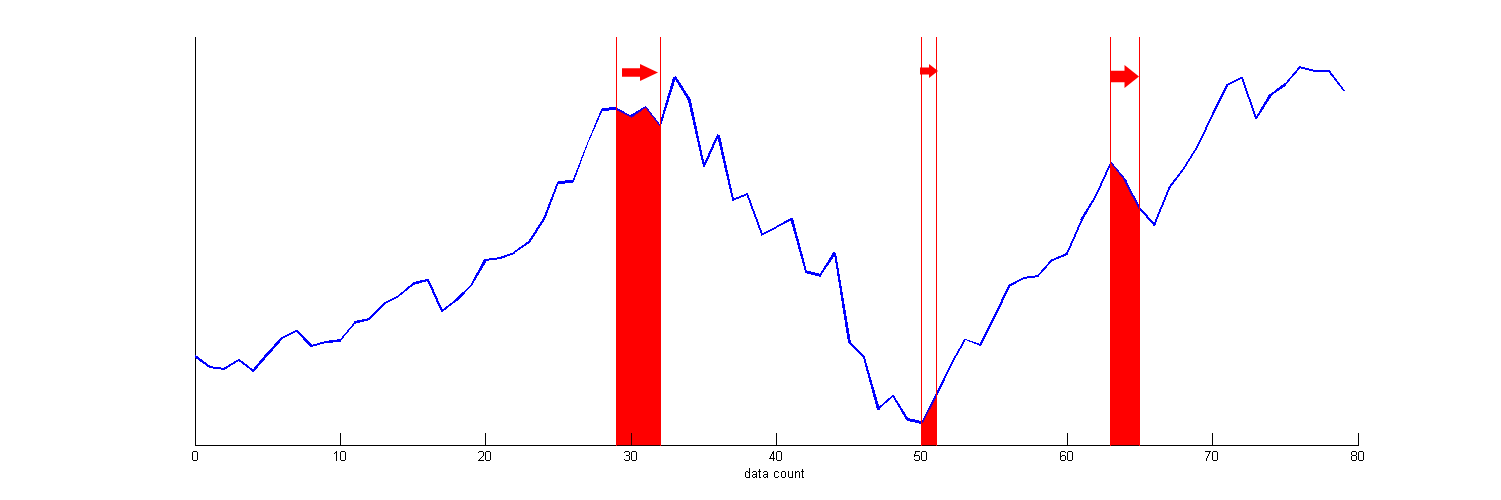}}
	\end{figure*}

For the comparison, the 3 most significant points detected by BFAST are 
compared to the 3 most significant points detected by FCPD, and the offset in 
the occurrence of the detection (measured in number of data points from the 
time domain) is reported. The offsets are the difference between each 
significant points from BFAST and FCPD (as illustrated in Fig. 
\ref{fig:comparison_illustration}).

\begin{figure*}[h]
	\captionsetup[subfigure]{position=b}
	\centering
	\caption[Histogram of the offsets of the 3 most significant segments of the 
	CICOP and the SWX data sets.]{Distribution (histogram) of the 
	offsets of the 3 most significant segments of the CICOP and the SWX 
	data sets. A negative offset indicates that the FCPD 
		algorithm detected the point before the BFAST algorithm.}
	\label{fig:comparison_BFAST_offset}
	\subcaptionbox{Distribution of the CICOP data set.}
	{\includegraphics[width=.48\linewidth]{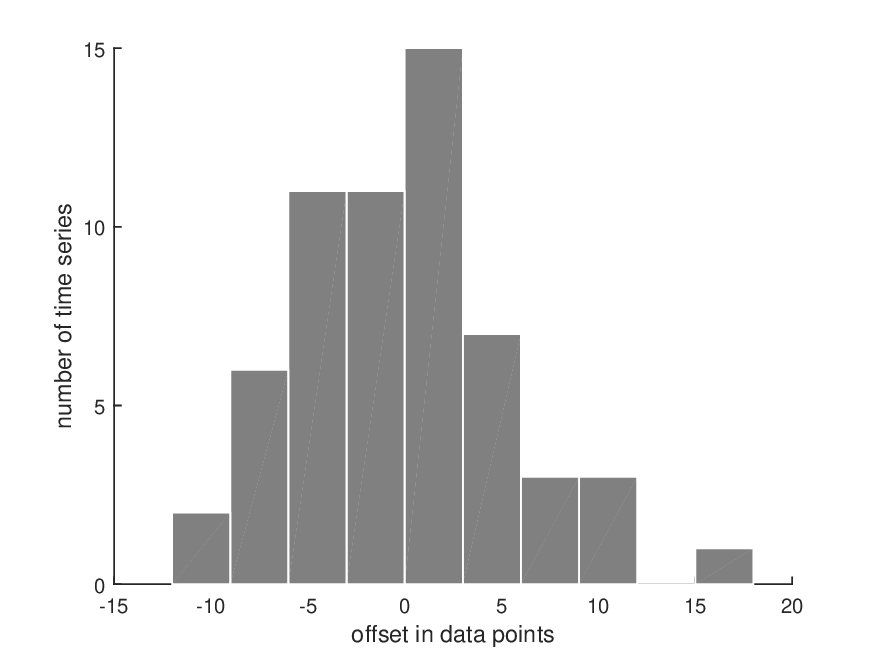}}
	\hfill
	\subcaptionbox{Distribution of the SWX data set.}
	{\includegraphics[width=.48\linewidth]{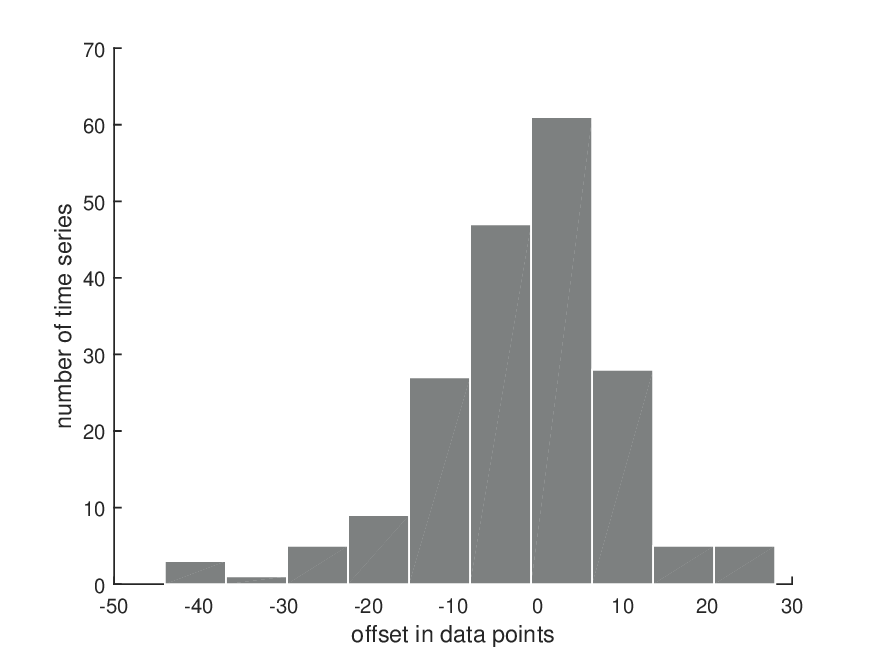}}
	\end{figure*}

For the 32 time series of the CICOP data set (the data set actually contains 60 
time series, but we only used the time series in which at least one change was 
detected by BFAST for a consistent comparison), the offsets are shown in Fig. 
\ref{fig:comparison_BFAST_offset} (left).

For the 64 time series of the SWX data set (the data set consists of 70 time 
series, but we only considered time series where all values were defined for 
the concerned period), the offsets are shown in Fig. 
\ref{fig:comparison_BFAST_offset} (right). 

Table \ref{tab:comparison_BFAST_offset} summarizes this comparison. It shows 
that the average offset for the CICOP data set is about $4.02$ data points, and 
about 
$8.15$ for the SWX data set. 

	\begin{table}[htp]
	\caption[Offset statistics for the 3 most significant changes compared with 
	BFAST.]{Offset statistics for the 3 most significant changes (O1, O2, 
	O3), compared with BFAST, in absolute values.}
	\label{tab:comparison_BFAST_offset}
	\centering
		\begin{tabular}{cccccc}
		\toprule[0.1em]
		& O1 & O2 & O3 & $\mu$ & $\sigma$
		\\  \cmidrule(r){2-4}
		CICOP & 4.66 & 3.89 & 3.50 & \textbf{4.02} & 0.48
		\\ 
		SWX & 7.09 & 5.88 & 11.49 & \textbf{8.15} & 2.41
		\\ \bottomrule[0.1em] 
		\end{tabular} 
	\end{table}

The average observed absolute offsets of $4$ data points for CICOP and $8$ data 
points for SWX with BFAST are considered as pretty good results in terms of 
similarity, especially when we know that FCPD is on-line and therefore only 
past values are used to detect 
change points, which is not the case with BFAST. Because of that, in the 
context of crime trends monitoring, BFAST cannot be used in a real environment. 
The distributions of the offsets from Figure 
\ref{fig:comparison_BFAST_offset} show that very few time series present an 
offset exceeding the absolute value of $10$ in data points, and also suggest 
that the offsets are equally distributed in terms of lag or of lead.

As an attempt to mimic the BFAST behavior and for comparison only, we used 
an automated method to fine-tune the FIS parameters, that is the settings of 
the membership functions, the linguistic variables and the rules. The 
MATLAB implementation of an adaptive-network-based fuzzy inference system 
(ANFIS, see \cite{Jang1993}) was used with both the CICOP and the SWX data set 
to compare the results from Table \ref{tab:comparison_BFAST_offset}, also with 
5 membership functions for the same inputs. A single FIS was trained with both 
data sets. The consequent average offsets are higher than the ``manual'' 
version, that is $7.00$ for CICOP and $11.72$ for SWX. 

This difficulty to extract more suitable parameters can have multiple causes. 
First, selecting the target of the ANFIS method (i.e., the supervised 
observations for the learning part) is far from obvious because many ways can 
be used to defined it. In our experiment, we decided to encode the score ``1'' 
when the discovered segment was within a region of $\pm 2$ data points of 
a BFAST detected point, and ``0'' otherwise. Second, both data set might not 
have enough observations for ANFIS to accurately learn the parameters from a 
machine learning perspective. And last, most of the ANFIS implementations only 
support Takagi-Sugeno inference types, as a result having less flexibility 
in the parameters and a different defuzzification method.  

Besides, FCPD presents a huge advantage in terms of complexity. For comparison 
only, the most computational consuming step of BFAST, that is, the detection of 
breaks based on \cite{Bai1994}, is of $O(N^2)$; whereas in 
FCPD, for the regression, the complexity is of $O(K^2)$, where $N$ is the 
number of observation and $K$ the degree of the regression ($K\ll N$). As 
illustration, the running time for the SWX data set is $42$ seconds for 
BFAST and $4$ seconds for FCPD on the same computer.

\subsection{Sensitivity analysis}
In this part the sensitivity of the proposed method in regard to its parameters 
is evaluated. The variation of the score of a query are observed with regard to 
changes in the parameters of the segmentation step (i.e., $K$, the degree; 
$\mathit{th_{DPU}}$ and $\mathit{th_{SSS}}$, the thresholds) and the querying 
step (i.e., the membership functions of the inputs). Default parameters are the 
ones used in the crime trends monitoring case study (Subsection 
\ref{ssec:ctm}). These variations are measured by computing the mean upper 
bound (i.e., the mean of the best 3 scores of the data set), the mean lower 
bound (i.e., the mean of the 
worst 3 scores of the data set), and the mean number of segments on both the 
CICOP and the SWX data sets.

	\begin{figure*}[htp]
	\captionsetup[subfigure]{position=b}
	\centering
	\caption[Sensitivity analysis of FCPD.]{(From top to bottom) Effects of a 
	change in the parameter $K$, 
	$\mathit{th_{DPU}}$, $\mathit{th_{SSS}}$; on the introduction of an offset 
	($\mathit{OFFSET}$); and on the input membership functions of the query 
	($\mathit{MF\_TYPE}$). The effects are measured on the mean lower bound and 
	the mean upper bound of the CICOP data set (left-hand) and the SWX data set 
	(right-hand). Lower bounds and upper bounds are calculated as the average 
	of the 3 worst/best scores. Score values are the mean of all time series of 
	the data set. The blue regions ($\mathit{K=5}$, $\mathit{th_{DPU}=0.05}$, 
	$\mathit{th_{SSS}=2}$, 	$\mathit{OFFSET=0}$, and $\mathit{MF\_TYPE=1}$ ) 
	are the reference values for the comparisons.}
	\label{fig:sensitivity_results}
	\includegraphics[width=.48\linewidth]{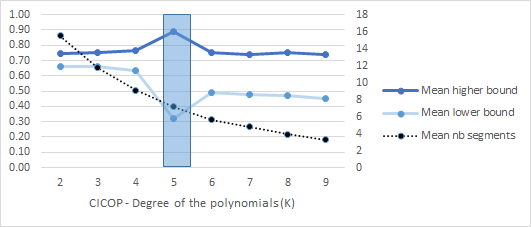}
	\hfill
	\includegraphics[width=.48\linewidth]{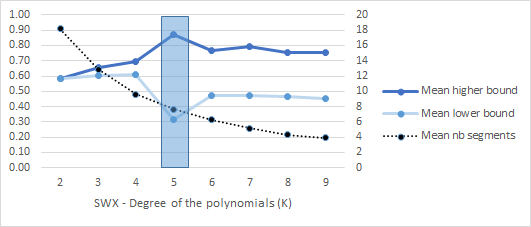}
	\hfill
	\includegraphics[width=.48\linewidth]{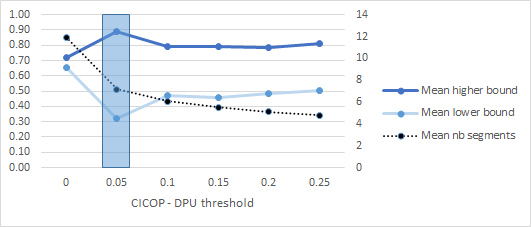}
	\hfill
	\includegraphics[width=.48\linewidth]{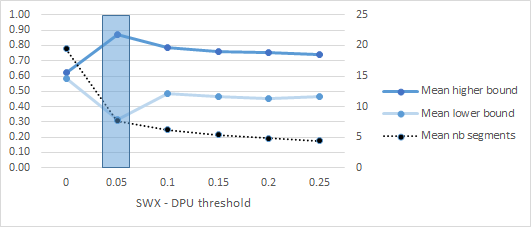}
	\hfill
	\includegraphics[width=.48\linewidth]{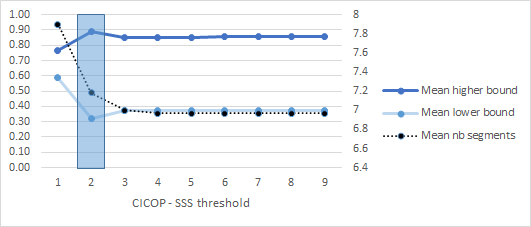}
	\hfill
	\includegraphics[width=.48\linewidth]{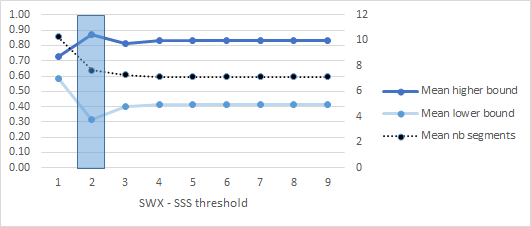}
	\hfill
	\includegraphics[width=.48\linewidth]{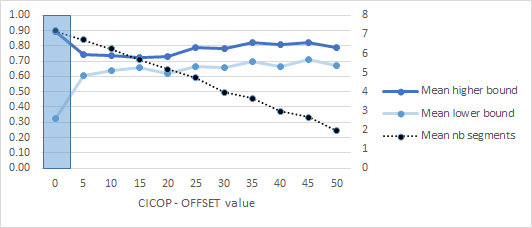}
	\hfill
	\includegraphics[width=.48\linewidth]{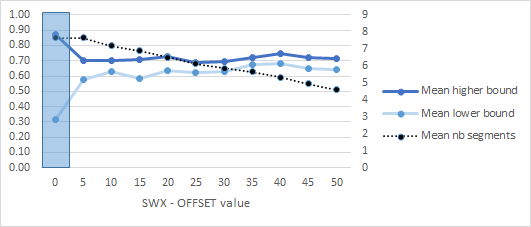}
	\hfill
	\includegraphics[width=.48\linewidth]{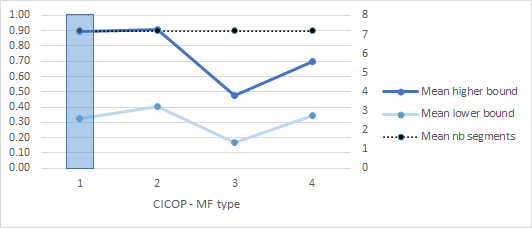}
	\hfill
	\includegraphics[width=.48\linewidth]{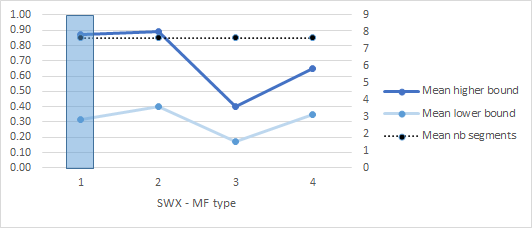}
	\end{figure*}

These singular changes are introduced either on the parameter K, either on the 
threshold $\mathit{th_{DPU}}$, either on the threshold $\mathit{th_{SSS}}$, 
either in the introduction of an offset in the x-axis, or in the input 
membership functions (Fig. \ref{fig:sensitivity_results}).	

\begin{figure*}[htp]
	\captionsetup[subfigure]{position=b}
	\centering
	\caption{Input membership functions used for the sensitivity analysis.}
	\label{fig:sensitivity_MF_type}
	\subcaptionbox{Set of membership functions denoted as $\mathit{TYPE\_1}$}
	{\includegraphics[width=.48\linewidth]{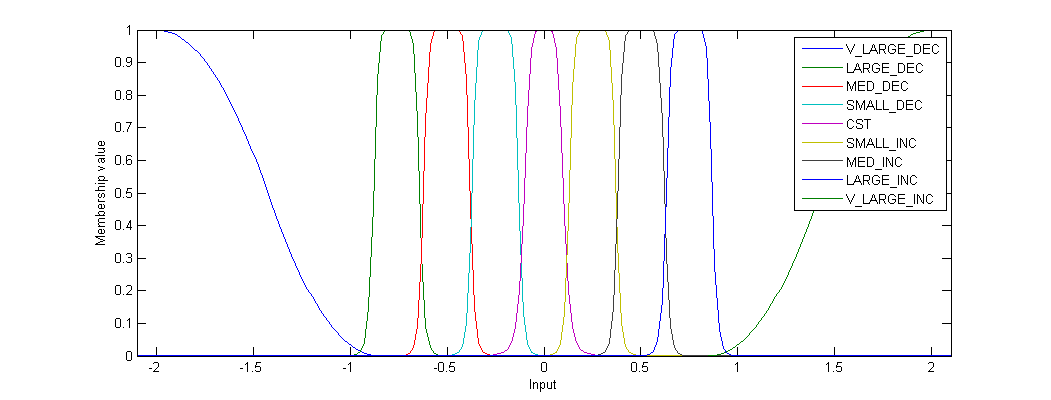}}
	\hfill
	\subcaptionbox{Set of membership functions denoted as $\mathit{TYPE\_2}$}
	{\includegraphics[width=.48\linewidth]{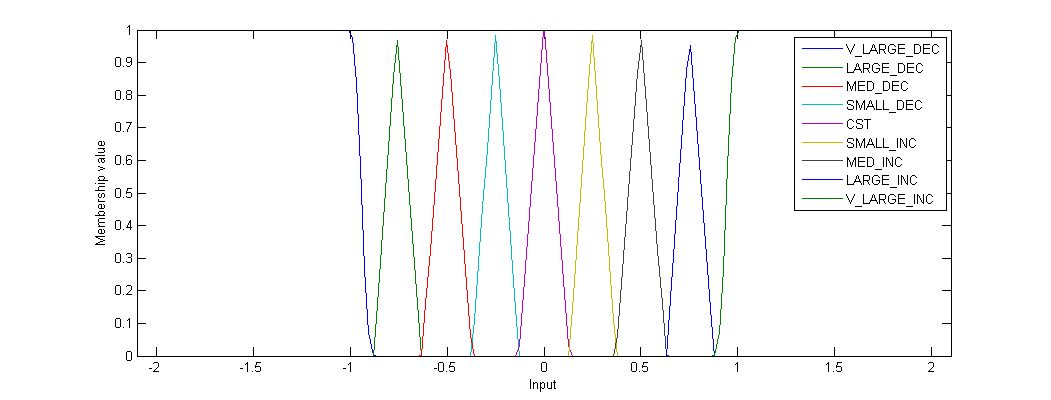}}
	\hfill
	\subcaptionbox{Set of membership functions denoted as $\mathit{TYPE\_3}$}
	{\includegraphics[width=.48\linewidth]{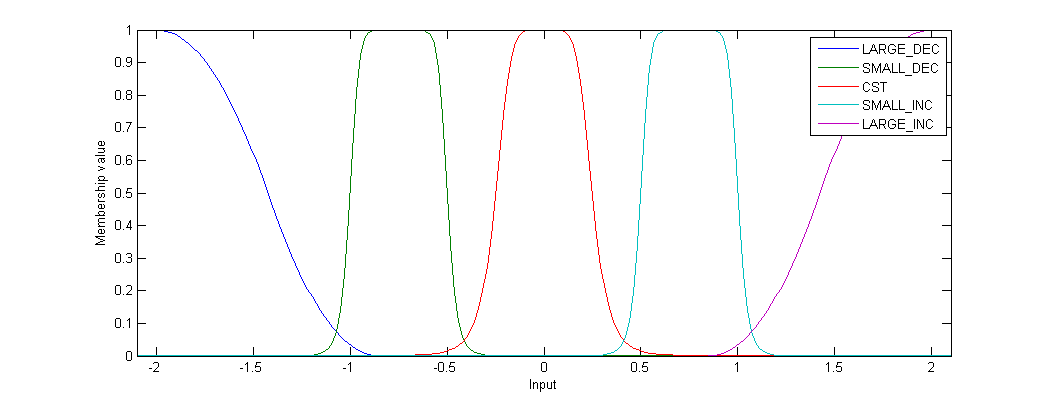}}
	\hfill
	\subcaptionbox{Set of membership functions denoted as $\mathit{TYPE\_4}$}
	{\includegraphics[width=.48\linewidth]{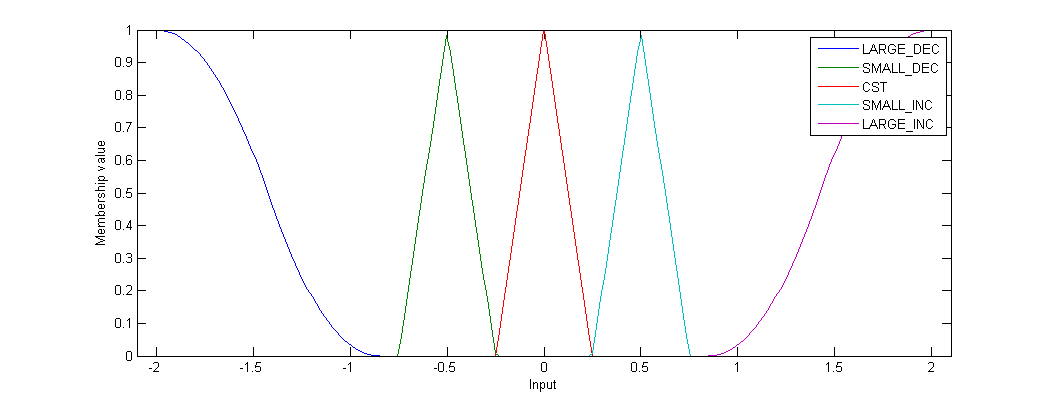}}
	\end{figure*}

The membership functions are depicted in Fig. \ref{fig:sensitivity_MF_type}. 
For membership functions $\mathit{TYPE\_3}$ and $\mathit{TYPE\_4}$, the rules 
have been consequently adapted (the number of membership having decreased, they 
need to be adapted according to the output variables): 

	\begin{enumerate}[a)]
	\item IF (\textit{var\_average} or \textit{var\_slope} is 
	\textit{large\_decrease}),
	THEN (\textit{score} is	\textit{very\_high})
	\item IF (\textit{var\_average} or \textit{var\_slope} is 
	\textit{small\_decrease}),
	THEN (\textit{score} is	\textit{medium})
	\item IF (\textit{var\_average} or \textit{var\_slope} is 
	\textit{constant}),
	THEN (\textit{score} is	\textit{very\_low})
	\item IF (\textit{var\_average} or \textit{var\_slope} is 
	\textit{small\_increase}),
	THEN (\textit{score} is	\textit{medium})
	\item IF (\textit{var\_average} or \textit{var\_slope} is 
	\textit{large\_increase}),
	THEN (\textit{score} is	\textit{very\_high})
	\end{enumerate}

	\begin{figure*}[htp]
		\captionsetup[subfigure]{position=b}
		\centering
		\caption[Interdependence between $K$ and the thresholds $thDPU$ and 
		$thSSS$.]{Interdependence between $K$ and the thresholds $thDPU$ and 
			$thSSS$. The scores are the mean for both the CICOP and the SWX 
			data 
			sets of the top 1 segment. Missing values indicate that no 
			segmentation 
			was found.}
		\label{fig:interdependence}
		\subcaptionbox{Interdependence between $K$ and $thDPU$.}
		{\includegraphics[width=.48\linewidth]{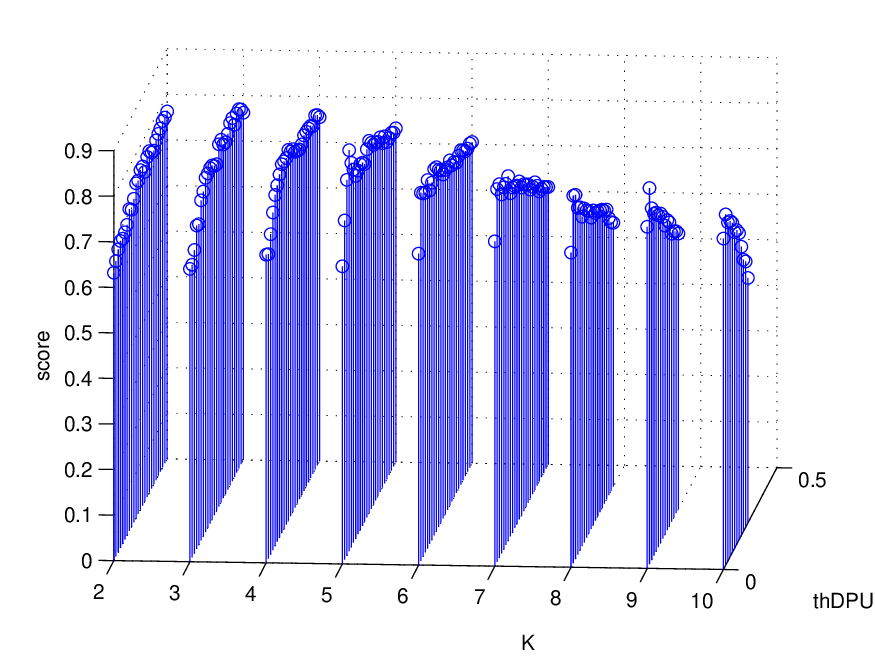}}
		\hfill
		\subcaptionbox{Interdependence between $K$ and $thSSS$.}
		{\includegraphics[width=.48\linewidth]{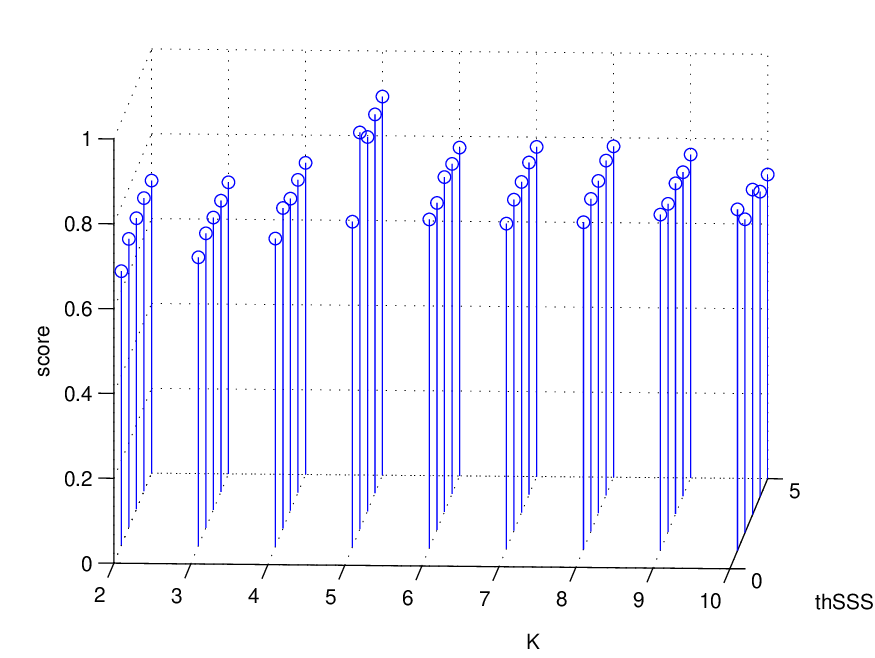}}
	\end{figure*} 
		
To better understand these results, let us take an example with a change of the 
degree on the CICOP data set (top-left of Fig.\ref{fig:sensitivity_results}). 
The reference value, denoted by the blue region, is shown as $\mathit{K=5}$, 
meaning that 3 best/worst scores are defined as \textit{reference} segments. 
Then, by modifying the value of $K$ only, the score of these 6 reference 
segments will be compared with their mean lower bounds and mean upper bound. 
More generally, we can interpret these measures by saying that the bigger the 
difference between the lower and upper bound is, the higher the method is 
sensitive to the considered parameter. The effect on the mean number of 
segments should also be taken into account.

The first interesting observation is that the method does not seem particularly 
sensitive to the change of a singular effect. Indeed, the difference between 
the mean lower and mean upper bounds are relatively constant in most 
settings and for both data sets. We however denote a slightly higher difference
with the thresholds.

Second, if we consider the mean upper bound, it remains 
high under most conditions, excepted for changes in the input membership 
functions. However, the mean lower bound seem to be pretty high. This could be 
explained by the mean number of segments, when it comes close to 6, i.e., the 
total number of change points considered only for the upper and lower bound.

Besides these singular changes, let us consider the interdependence of the 
parameters, that is between the segmentation step and the query step. 
The parameter $K$ and the segmentation thresholds impact the
segmentation results. If more segments are considered (i.e., by decreasing $K$
or the thresholds, as seen in Fig. \ref{fig:sensitivity_results}), 
the coefficients will describe the segments more precisely, but very local 
changes will be reported with respect to the query. On the other hand, settings 
parameters that create fewer segments (i.e., by increasing $K$ or the 
thresholds), the coefficients will not be able to precisely describe the 
segment, resulting in inappropriate changes reported with respect to the query. 
Interdependence between the degree $K$ and the thresholds $thDPU$ and $thSSS$ 
are shown in Fig. \ref{fig:interdependence}. The score is 
computed as the mean of the top 1 segment for both the CICOP and SWX data set. 
The variability seems to be relatively low, in the sense that modifying one 
parameter does not have a marked effect on the other.

\section{Discussion}
\label{sec:discussion}
As mentioned in the previous sections, the proposed method presents three main 
advantages: (a) an intuitive and meaningful representation of the time series, 
(b) a dynamic and on-line segmentation method, and (c) a flexible and 
understandable querying system.

These claims are supported by our experiments: the two case studies 
illustrate the flexibility and the feasibility of the intuitive querying of 
change points; the comparison with the BFAST algorithm show similar results in 
terms of accuracy; the sensibility analysis show that the parameters of the 
segmentation part can be consistently determined; and in terms of 
computational complexity, FCPD is much more efficient than BFAST.


\section{Conclusions}
\label{sec:conclusion}
A method for the detection of change points within 
crime-related time series was described and tested with different data sets. 
The combination of a meaningful representation, a dynamic 
segmentation, and a fuzzy inference system delivers the possibility, even for 
experts not related to data mining, to intuitively find change points by 
describing geometric properties in linguistic terms. More broadly, the 
considerable flexibility of the method makes possible the 
use of the method in any application domain, with a great potential 
in crime analysis.

Future work suggest further investigation on the use of mining methods to 
automatically discover the most appropriate membership functions of the 
inference system in order to mimic the behavior of existing algorithms. This 
alternative could present a gain in the accuracy of the detected change points, 
however, the opposing view is a loss in the understanding of the inference 
system. 

Also, an implementation of a crime trends monitoring process in a real 
environment should be tested and the results assessed in real time by crime 
analysts.

\section*{Acknowledgements}
\label{acknowledgements}
The authors are grateful to the Swiss National Science Foundation (SNSF) for 
the support of this work under project no. 156287. The authors would also like 
to thank the Police de Sûreté du Canton de Vaud for the access provided to the 
data, and particularly Sylvain Ioset and Damien Dessimoz for their support.

\section*{References}
\bibliographystyle{elsarticle-num} 
\bibliography{article}

\begin{thebibliography}{10}
\expandafter\ifx\csname url\endcsname\relax
  \def\url#1{\texttt{#1}}\fi
\expandafter\ifx\csname urlprefix\endcsname\relax\def\urlprefix{URL }\fi
\expandafter\ifx\csname href\endcsname\relax
  \def\href#1#2{#2} \def\path#1{#1}\fi

\bibitem{Albertetti2012}
F.~Albertetti, K.~Stoffel, From police reports to datamarts: Towards a crime
  analysis framework, in: S.~Srihari, K.~Franke (Eds.), Proceedings of the 5th
  International Workshop, IWCF 2012, Tsukuba, Japan, 2012, pp. 48--59.

\bibitem{Albertetti2013}
F.~Albertetti, P.~Cotofrei, L.~Grossrieder, O.~Ribaux, K.~Stoffel, Crime
  linkage: A fuzzy mcdm approach, in: Intelligence and Security Informatics
  (ISI), 2013 IEEE International Conference on, 2013, pp. 1--3.
\newblock \href {https://doi.org/10.1109/ISI.2013.6578772}
  {\path{doi:10.1109/ISI.2013.6578772}}.

\bibitem{Albertetti2013b}
F.~Albertetti, P.~Cotofrei, L.~Grossrieder, O.~Ribaux, K.~Stoffel, The crilim
  methodology: Crime linkage with a fuzzy mcdm approach, in: Intelligence and
  Security Informatics Conference (EISIC), 2013 European, 2013, pp. 67--74.
\newblock \href {https://doi.org/10.1109/EISIC.2013.17}
  {\path{doi:10.1109/EISIC.2013.17}}.

\bibitem{Grossrieder2013}
L.~Grossrieder, F.~Albertetti, K.~Stoffel, O.~Ribaux, Des données aux
  connaissances, un chemin difficile: réflexion sur la place du data mining en
  analyse criminelle, Revue Internationale de Criminologie et de Police
  Technique et Scientifique 66 (2013) 99--116.

\bibitem{Boba2009}
R.~Boba, Crime Analysis with Crime Mapping, Thousand Oaks, CA: Sage, 2009.

\bibitem{Felson1998}
M.~Felson, R.~V. Clarke, Opportunity makes the thief, Police Research Series,
  Paper 98 (1998).

\bibitem{Fu2011}
T.-c. Fu, A review on time series data mining, Engineering Applications of
  Artificial Intelligence 24~(1) (2011) 164--181.

\bibitem{Last2001}
M.~Last, Y.~Klein, A.~Kandel, Knowledge discovery in time series databases,
  Systems, Man, and Cybernetics, Part B: Cybernetics, IEEE Transactions on
  31~(1) (2001) 160--169.

\bibitem{Lin2003}
J.~Lin, E.~Keogh, S.~Lonardi, B.~Chiu, A symbolic representation of time
  series, with implications for streaming algorithms, in: Proceedings of the
  8th ACM SIGMOD workshop on Research issues in data mining and knowledge
  discovery, ACM, 2003, pp. 2--11.

\bibitem{Lin2007}
J.~Lin, E.~Keogh, L.~Wei, S.~Lonardi, Experiencing sax: a novel symbolic
  representation of time series, Data Mining and Knowledge Discovery 15~(2)
  (2007) 107--144.

\bibitem{Fuchs2010b}
E.~Fuchs, T.~Gruber, H.~Pree, B.~Sick, Temporal data mining using shape space
  representations of time series, Neurocomputing 74~(1) (2010) 379--393.

\bibitem{Keogh2001}
E.~J. Keogh, S.~Chu, D.~M. Hart, M.~J. Pazzani, An online algorithm for
  segmenting time series, in: ICDM, 2001, pp. 289--296.
\newblock \href {https://doi.org/10.1109/ICDM.2001.989531}
  {\path{doi:10.1109/ICDM.2001.989531}}.

\bibitem{Keogh2004}
E.~Keogh, S.~Chu, D.~Hart, M.~Pazzani, Segmenting time series: A survey and
  novel approach, Data mining in time series databases 57 (2004) 1--22.

\bibitem{Fuchs2010}
E.~Fuchs, T.~Gruber, J.~Nitschke, B.~Sick, Online segmentation of time series
  based on polynomial least-squares approximations, Pattern Analysis and
  Machine Intelligence, IEEE Transactions on 32~(12) (2010) 2232--2245.

\bibitem{Chung2002}
F.-l. Chung, T.-c. Fu, R.~Luk, V.~Ng, Evolutionary time series segmentation for
  stock data mining, in: Data Mining, 2002. ICDM 2003. Proceedings. 2002 IEEE
  International Conference on, IEEE, 2002, pp. 83--90.

\bibitem{Song1993}
Q.~Song, B.~S. Chissom, Forecasting enrollments with fuzzy time series---part
  i, Fuzzy sets and systems 54~(1) (1993) 1--9.

\bibitem{Song1993b}
Q.~Song, B.~S. Chissom, Fuzzy time series and its models, Fuzzy sets and
  systems 54~(3) (1993) 269--277.

\bibitem{Chen1996}
S.-M. Chen, Forecasting enrollments based on fuzzy time series, Fuzzy sets and
  systems 81~(3) (1996) 311--319.

\bibitem{Hwang1998}
J.-R. Hwang, S.-M. Chen, C.-H. Lee, Handling forecasting problems using fuzzy
  time series, Fuzzy sets and systems 100~(1) (1998) 217--228.

\bibitem{Huarng2001}
K.~Huarng, Heuristic models of fuzzy time series for forecasting, Fuzzy sets
  and systems 123~(3) (2001) 369--386.

\bibitem{Chen2012}
C.-H. Chen, T.-P. Hong, V.~S. Tseng, Fuzzy data mining for time-series data,
  Applied Soft Computing 12~(1) (2012) 536--542.

\bibitem{Mamdani1974}
E.~Mamdani, Application of fuzzy algorithms for control of simple dynamic
  plant, Proceedings of the Institution of Electrical Engineers 121~(12) (1974)
  1585--1588.
\newblock \href {https://doi.org/10.1049/piee.1974.0328}
  {\path{doi:10.1049/piee.1974.0328}}.

\bibitem{Sugeno1985}
M.~Sugeno, T.~Takagi, Fuzzy identification of systems and its applications to
  modeling and control, IEEE Transactions On Systems Man And Cybernetics 15~(1)
  (1985) 116--132.

\bibitem{Lee2006}
C.-H.~L. Lee, A.~Liu, W.-S. Chen, Pattern discovery of fuzzy time series for
  financial prediction, Knowledge and Data Engineering, IEEE Transactions on
  18~(5) (2006) 613--625.

\bibitem{Guener2014}
H.~A.~A. Güner, H.~A. Yumuk, Application of a fuzzy inference system for the
  prediction of longshore sediment transport, Applied Ocean Research 48 (2014)
  162--175.

\bibitem{Jang1993}
J.~Jang, Anfis - adaptive-network-based fuzzy inference system, Systems, Man
  and Cybernetics, IEEE Transactions on 23~(3) (1993) 665--685.
\newblock \href {https://doi.org/10.1109/21.256541}
  {\path{doi:10.1109/21.256541}}.

\bibitem{Song2000}
Q.~Song, N.~Kasabov, Dynamic evolving neuro-fuzzy inference system (denfis):
  On-line learning and application for time-series prediction, in: Proc. 6th
  International Conference on Soft Computing, Citeseer, 2000, pp. 696--701.

\bibitem{Gueler2005}
I.~Güler, E.~D. Übeyli, Adaptive neuro-fuzzy inference system for
  classification of eeg signals using wavelet coefficients, Journal of
  neuroscience methods 148~(2) (2005) 113--121.

\bibitem{Zounemat-Kermani2008}
M.~Zounemat-Kermani, M.~Teshnehlab, Using adaptive neuro-fuzzy inference system
  for hydrological time series prediction, Applied Soft Computing 8~(2) (2008)
  928--936.

\bibitem{Basseville1993}
M.~Basseville, I.~V. Nikiforov, et~al., Detection of abrupt changes: theory and
  application, Vol. 104, Prentice Hall Englewood Cliffs, 1993.

\bibitem{Reeves2007}
J.~Reeves, J.~Chen, X.~L. Wang, R.~Lund, Q.~Q. Lu, A review and comparison of
  changepoint detection techniques for climate data, Journal of Applied
  Meteorology and Climatology 46~(6) (2007) 900--915.

\bibitem{Wu1999}
B.~Wu, M.-H. Chen, Use of fuzzy statistical technique in change periods
  detection of nonlinear time series, Applied Mathematics and Computation
  99~(2) (1999) 241--254.

\bibitem{Kumar2001}
K.~Kumar, B.~Wu, Detection of change points in time series analysis with fuzzy
  statistics, International Journal of Systems Science 32~(9) (2001)
  1185--1192.
\newblock \href
  {http://arxiv.org/abs/http://dx.doi.org/10.1080/00207720110034698}
  {\path{arXiv:http://dx.doi.org/10.1080/00207720110034698}}, \href
  {https://doi.org/10.1080/00207720110034698}
  {\path{doi:10.1080/00207720110034698}}.

\bibitem{Verbesselt2010}
J.~Verbesselt, R.~Hyndman, G.~Newnham, D.~Culvenor, Detecting trend and
  seasonal changes in satellite image time series, Remote sensing of
  Environment 114~(1) (2010) 106--115.

\bibitem{Verbesselt2010b}
J.~Verbesselt, R.~Hyndman, A.~Zeileis, D.~Culvenor, Phenological change
  detection while accounting for abrupt and gradual trends in satellite image
  time series, Remote Sensing of Environment 114~(12) (2010) 2970--2980.

\bibitem{Cappelli2013}
C.~Cappelli, P.~D'Urso, F.~D. Iorio, Change point analysis of imprecise time
  series, Fuzzy Sets and Systems 225~(0) (2013) 23--38, theme: Fuzzy Systems.
\newblock \href {https://doi.org/10.1016/j.fss.2013.03.001}
  {\path{doi:10.1016/j.fss.2013.03.001}}.

\bibitem{Chen2013}
X.~C. Chen, K.~Steinhaeuser, S.~Boriah, S.~Chatterjee, V.~Kumar, Contextual
  time series change detection., in: SDM, SIAM, 2013, pp. 503--511.

\bibitem{Yamanishi2002}
K.~Yamanishi, J.-i. Takeuchi, A unifying framework for detecting outliers and
  change points from non-stationary time series data, in: Proceedings of the
  eighth ACM SIGKDD international conference on Knowledge discovery and data
  mining, ACM, 2002, pp. 676--681.

\bibitem{Takeuchi2006}
J.-i. Takeuchi, K.~Yamanishi, A unifying framework for detecting outliers and
  change points from time series, Knowledge and Data Engineering, IEEE
  Transactions on 18~(4) (2006) 482--492.

\bibitem{Moreno-Garcia2014}
J.~Moreno-Garcia, L.~Rodriguez-Benitez, J.~Giralt, E.~del Castillo, The
  generation of qualitative descriptions of multivariate time series using
  fuzzy logic, Applied Soft Computing 23~(0) (2014) 546 -- 555.
\newblock \href {https://doi.org/http://dx.doi.org/10.1016/j.asoc.2014.05.021}
  {\path{doi:http://dx.doi.org/10.1016/j.asoc.2014.05.021}}.

\bibitem{Yu2001}
J.-R. Yu, G.-H. Tzeng, H.-L. Li, General fuzzy piecewise regression analysis
  with automatic change-point detection, Fuzzy Sets and Systems 119~(2) (2001)
  247 -- 257.
\newblock \href
  {https://doi.org/http://dx.doi.org/10.1016/S0165-0114(98)00384-4}
  {\path{doi:http://dx.doi.org/10.1016/S0165-0114(98)00384-4}}.

\bibitem{Wang2004b}
H.~Wang, D.~Zhang, K.~G. Shin, Change-point monitoring for the detection of dos
  attacks, IEEE Transactions on Dependable and Secure Computing 1~(4) (2004)
  193--208, copyright - Copyright IEEE Computer Society Oct-Dec 2004;
  Caractéristique du document - references; Dernière mise à jour -
  2010-06-07.

\bibitem{Bai1994}
J.~Bai, Least squares estimation of a shift in linear processes, Journal of
  Time Series Analysis 15~(5) (1994) 453--472.

\end{thebibliography}

\section*{} 
\setlength\intextsep{0pt} 

\newpage
\subsection*{}
	\begin{wrapfigure}{l}{0.12\textwidth}
	\includegraphics[width=27mm]{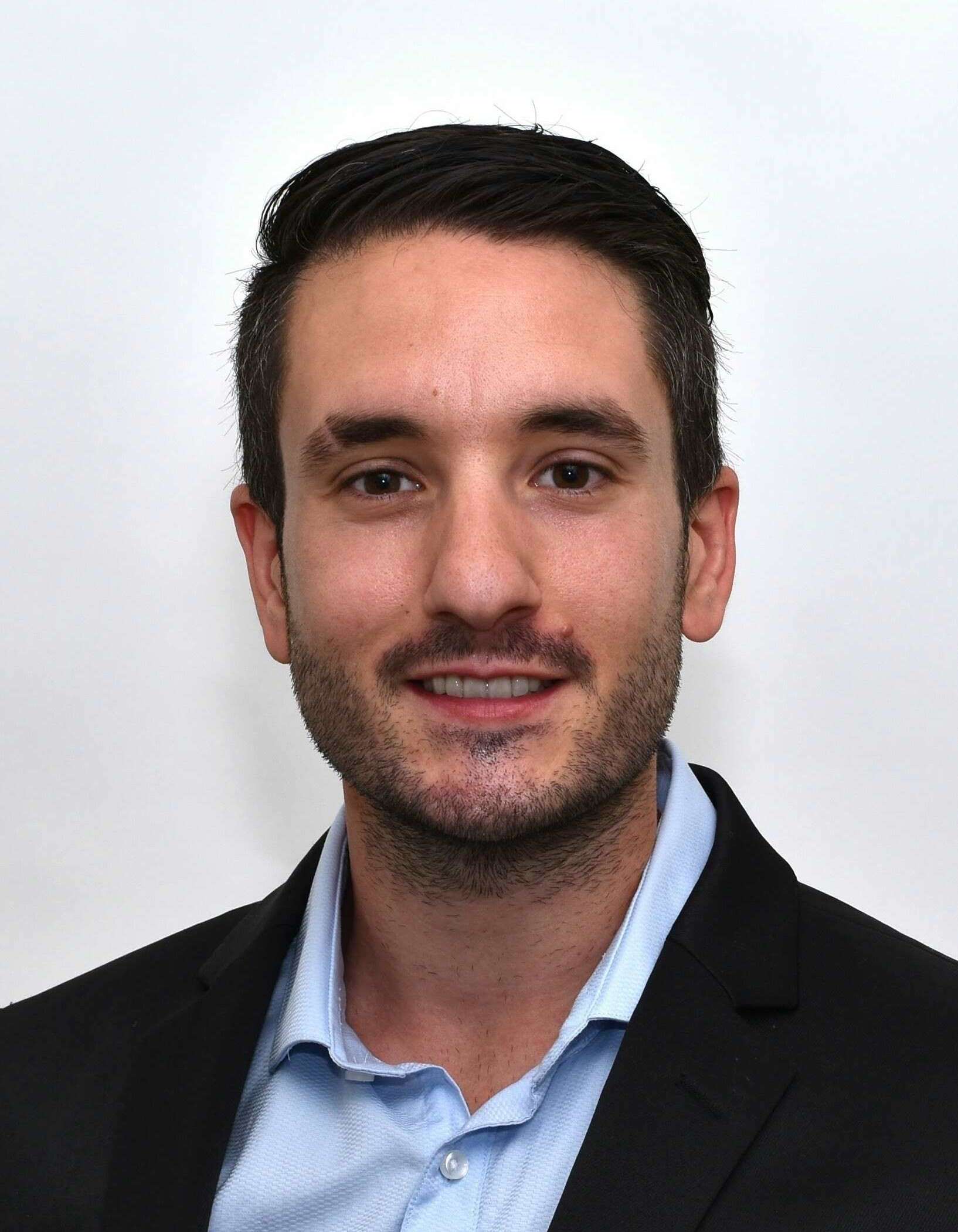}
	\end{wrapfigure}
\nopagebreak \noindent
\textbf{Fabrizio Albertetti} received his engineer's degree in 
computer science in 2009 and his master's degree in information systems in 
2011. He is currently holding a PhD candidate and teaching assistant position 
at the University of Neuchatel (Switzerland), with an interest in developing 
knowledge extraction methods for crime analysis. He is also involved in an 
SNSF interdisciplinary project on the application of computational methods in 
crime analysis.

\subsection*{}
	\begin{wrapfigure}{l}{0.12\textwidth}
	\includegraphics[width=27mm]{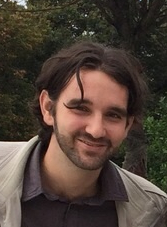}
	\end{wrapfigure}
\nopagebreak \noindent
\textbf{Lionel Grossrieder} is a SNSF Doctoral Student in the School 
of Criminal Justice of Lausanne, Switzerland. He received his bachelor degree 
in psychology in 2009 and his master degree in criminology in 2011. His 
research interests include crime analysis, environmental criminology and 
forensic intelligence. He is currently involved in an interdisciplinary 
project 
on application of computational methods in crime analysis.

\subsection*{}
	\begin{wrapfigure}{l}{0.12\textwidth}
	\includegraphics[width=27mm]{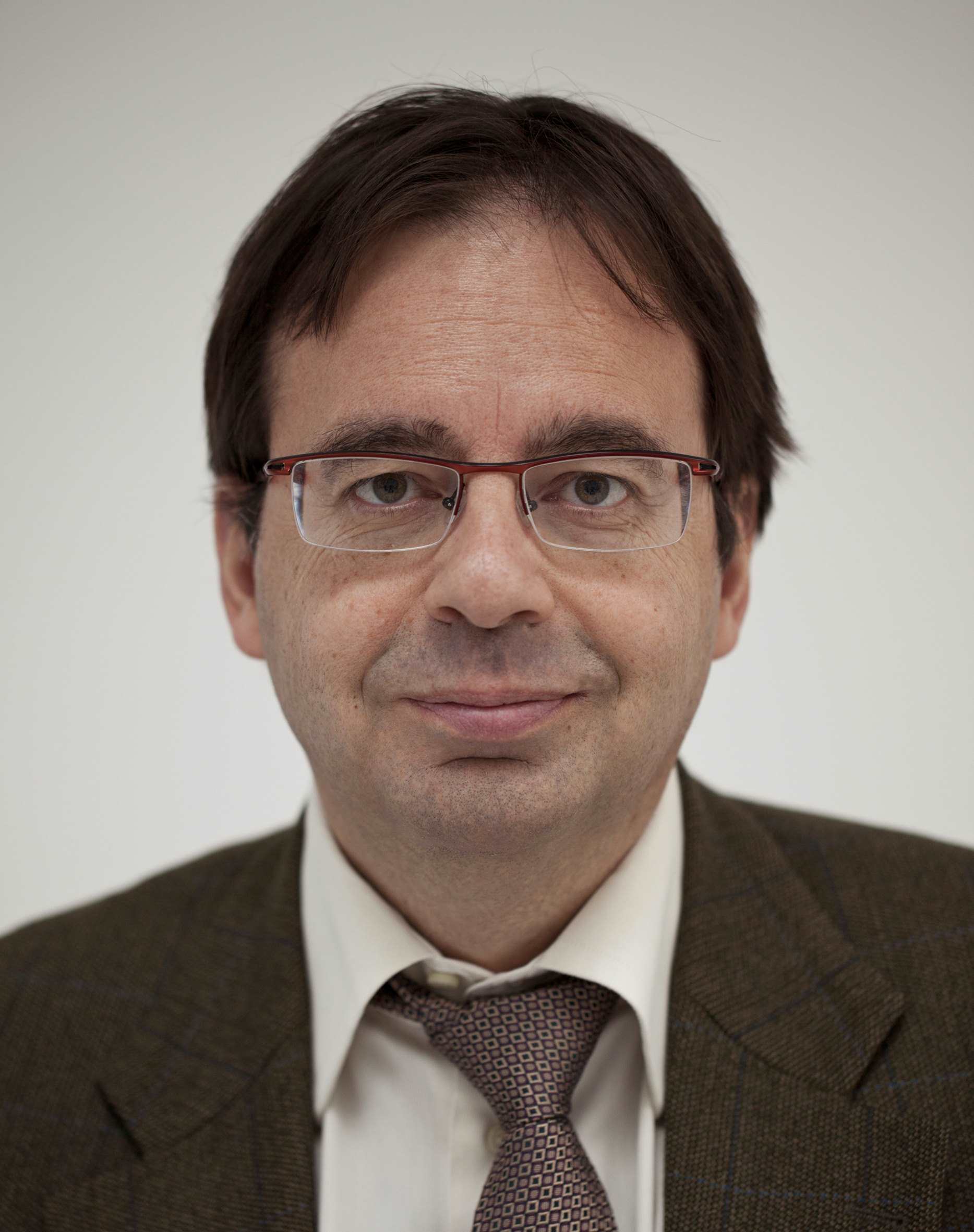}
	\end{wrapfigure}
\nopagebreak \noindent
\textbf{Olivier Ribaux} is the director of the school of criminal 
justice, University of Lausanne, Switzerland. After his 10 years career as a 
crime analyst at a Swiss canton’s police, he joined the University to conduct 
researches on the contribution of forensic science to intelligence-led 
policing. His studies range from the formalization of methods to the 
development of computerized tools, mostly for the analysis of repetitive 
crimes. He trains students, law enforcement officers and magistrates in crime 
analysis and forensic intelligence. He is the author of many peer-reviewed 
scientific publications, book chapters and of a book on forensic intelligence 
(in French).


\subsection*{}
	\begin{wrapfigure}{l}{0.12\textwidth}
	\includegraphics[width=27mm]{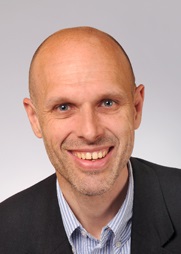}
	\end{wrapfigure}

\nopagebreak \noindent
\textbf{Kilian Stoffel} is a professor of computer science at the 
University of Neuchatel. He has received a master’s degree in mathematics and 
in computer science and he holds a PhD in computer science of the University 
of Fribourg (Switzerland). Before joining the Faculty of the University of 
Neuchatel, he was with by the University of Maryland at College Park and by 
Johns Hopkins University. His main interests of research cover the domains of 
Knowledge Discovery and Knowledge Representation. He has authored and/or 
co-authored over seventy papers in the domain.


\end{document}